\documentclass{article}
\PassOptionsToPackage{table,xcdraw}{xcolor}
\usepackage{microtype}
\usepackage{graphicx}
\usepackage{subfigure}
\usepackage{booktabs} % for professional tables
\usepackage{pifont}

\usepackage{hyperref}

\usepackage[accepted]{icml2025}

\usepackage{amsmath}
\usepackage{amssymb}
\usepackage{mathtools}
\usepackage{amsthm}
\usepackage{enumitem}
\usepackage{booktabs}
\usepackage{multirow}
\usepackage{graphicx}
\usepackage{xcolor}
\usepackage{float}

\usepackage[capitalize,noabbrev]{cleveref}

\newcommand{\bc}{\texttt{BetaConform}}

\theoremstyle{plain}
\newtheorem{theorem}{Theorem}[section]
\newtheorem{proposition}[theorem]{Proposition}

\newtheorem{corollary}[theorem]{Corollary}
\theoremstyle{definition}
\newtheorem{definition}[theorem]{Definition}
\newtheorem{assumption}[theorem]{Assumption}
\theoremstyle{remark}

\usepackage[textsize=tiny]{todonotes}

\icmltitlerunning{Efficient MAP Estimation of LLM Judgment Performance with Prior Transfer}

\begin{document}

\twocolumn[
\icmltitle{Efficient MAP Estimation of LLM Judgment Performance with Prior Transfer}



\icmlsetsymbol{equal}{*}

\begin{icmlauthorlist}
\icmlauthor{Huaizhi Qu}{UNC}
\icmlauthor{Inyoung Choi}{UPenn}
\icmlauthor{Zhen Tan}{ASU}
\icmlauthor{Song Wang}{UVA}
\icmlauthor{Sukwon Yun}{UNC}
\icmlauthor{Qi Long}{UPenn}
\icmlauthor{Faizan Siddiqui}{HRI}
\icmlauthor{Kwonjoon Lee}{HRI}
\icmlauthor{Tianlong Chen}{UNC}
\end{icmlauthorlist}

\icmlaffiliation{UNC}{The University of North Carolina at Chapel Hill}
\icmlaffiliation{HRI}{Honda Research Institute USA}
\icmlaffiliation{ASU}{Arizona State University}
\icmlaffiliation{UVA}{University of Virginia}
\icmlaffiliation{UPenn}{University of Pennsylvania}

\icmlcorrespondingauthor{Tianlong Chen}{tianlong@cs.unc.edu}
\icmlcorrespondingauthor{Kwonjoon Lee}{kwonjoon\_lee@honda-ri.com}

\icmlkeywords{Machine Learning, ICML}

\vskip 0.3in
]



\printAffiliationsAndNotice{} % otherwise use the standard text.

\begin{abstract}

LLM ensembles are widely used for LLM judges. However, how to estimate their accuracy, especially in an efficient way, is unknown. In this paper, we present a principled \textit{maximum a posteriori} (MAP) framework for an economical and precise estimation of the performance of LLM ensemble judgment. We first propose a mixture of Beta-Binomial distributions to model the judgment distribution, revising from the vanilla Binomial distribution. Next, we introduce a conformal prediction-driven approach that enables adaptive stopping during iterative sampling to balance accuracy with efficiency. Furthermore, we design a prior transfer mechanism that utilizes learned distributions on open-source datasets to improve estimation on a target dataset when only scarce annotations are available. Finally, we present \bc, a framework that integrates our distribution assumption, adaptive stopping, and the prior transfer mechanism to deliver a theoretically guaranteed distribution estimation of LLM ensemble judgment with minimum labeled samples. \bc\ is also validated empirically. For instance, with only $10$ samples from the TruthfulQA dataset, for a Llama ensembled judge, \bc\ gauges its performance with error margin as small as $3.37\%$.
\vspace{-5mm}
\end{abstract}

\section{Introduction}
\label{sec:intro}

With the improving performance of large language models (LLMs), there is a proliferation of adopting LLMs as judges for various tasks \cite{liang2023encouraging,yuan2024self,zhang2025lessonsdevelopingprocessreward}. In applications of LLM judge ensembles, the judgment distribution is critical to the service quality \cite{chen2024seeing,schoenegger2024wisdomsiliconcrowdllm,qiu2025ensemblelargelanguagemodels}. Many datasets \cite{zheng2023judging,zeng2023evaluating,yuan2024r} have been employed to evaluate the performance of LLM judges. However, these datasets rely on human annotations, which are impractical at large scale due to the substantial time and financial costs of annotating. This challenge highlights the need of \textit{how to estimate the LLM ensemble judging performance efficiently}.

\begin{figure}[t]
\includegraphics[width=1\columnwidth]{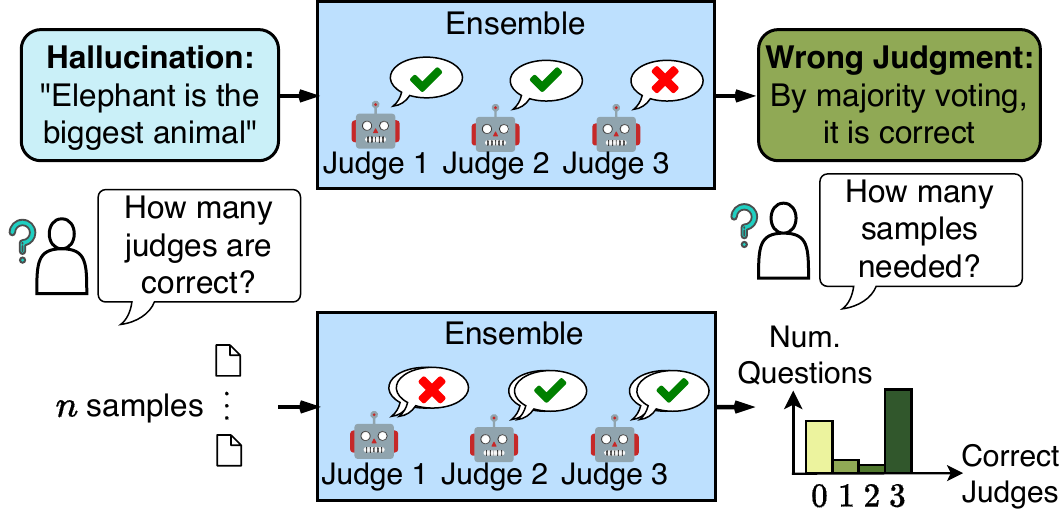}
\vspace{-7mm}
\caption{In this paper, we aim to answer $(1)$ how to estimate the judgment distribution of LLM ensemble on a dataset, and $(2)$ how to achieve efficient estimation to reduce annotation effort.}
\vspace{-3mm}
\label{fig:teaser}
\end{figure}

In this work, we consider judgment distribution estimation:
\begin{equation*}
    \mathbb{P}(\text{\# correct judgments}=n \mid k \text{ LLMs judge sample}\ x).
\end{equation*}
We propose an efficient method for MAP estimation of the distribution of LLM ensemble judgment to answer two research questions shown in Figure \ref{fig:teaser}. 
\begin{enumerate}[noitemsep, topsep=0pt]
    \item \textbf{RQ1:} How to estimate the judgment distribution? \label{rq1}
    \item \textbf{RQ2:} How many samples are needed for estimation? \label{rq2}
\end{enumerate}
Given a small number of samples, one intuitive estimation is to directly adopt the distribution of the samples as the judgment distribution on the entire dataset. However, this is susceptible to the sampling bias. To avoid this, one common practice is to first calculate the single LLM accuracy on the samples and then model the distribution on the full dataset as Binomial. We first posit that the judgment distribution is not Binomial. Theoretically, a Binomial distribution implies increasing accuracy in majority voting as the ensemble size grows \cite{de2014essai,austen1996information}. However, this is unrealistic since the accuracy of LLM ensembles remains bounded even with a large number of judges. To testify to this, we start by observing the distribution of LLM ensemble judges on various benchmarks. We find marked deviations from the Binomial distribution and show a stratification between questions that can be classified as ``easy'' and ``hard''. To this end, we propose to model the judgment distribution with a mixture of Beta-Binomial distributions to reflect the stratification.
% : one corresponds to the "easy" questions that LLMs tend to keep correct, and one distribution reflects the questions that LLMs are consistently wrong.
We show that under this assumption, by utilizing an expectation maximization (EM) estimation method, it can achieve accurate judgment distribution estimation with high data efficiency.

To rigorously guide the sampling process and determine how many samples to use for the estimation, we draw inspiration from the conformal prediction (CP) \cite{shafer2008tutorial,fontana2023conformal} that can efficiently estimate the sampling deviation. Based on this, we propose a novel adaptive stopping strategy for iterative sampling, designed to meet a pre-defined deviation threshold. Our experiments demonstrate the effectiveness of this method for limiting the sample amount while maintaining high estimation precision.

Moreover, we hypothesize that the prior knowledge of judgment distribution on open-source datasets can benefit the estimation of a new dataset when only a few samples are available. To achieve this, we propose a text similarity-based distribution prior transfer mechanism. This method embeds text inputs from both source and target datasets and calculates embedding similarities to determine the transfer weight. Our design greatly improves the estimation accuracy when transferring from similar datasets and avoids performance degradation when the datasets are distinct. Notably, this method relies solely on the text inputs, making it practical for application to vast amounts of unlabeled data.

Our contribution can be summarized as follows:
\begin{itemize}[noitemsep, topsep=0pt]
    \item We present pioneering work in judgment distribution estimation. We point out that the Binomial assumption of judgment distribution is inaccurate. By replacing it with a mixture of Beta-Binomial distributions, we could achieve efficient and accurate estimation.
    \item We design a rigorous conformal prediction-based adaptive stopping strategy during iterative sampling when the sampling deviation is sufficiently low.
    \item We introduce a distribution prior transfer mechanism that leverages judgment distributions on open-source datasets to improve few-sample estimations.
    \item Extensive experiments show \bc's high estimation efficiency. For example, using only $10$ samples could result in an average of $10.84\%$ error margin.
\end{itemize}

\section{Related Works}
\paragraph{LLMs for Judgment.} 
% The evaluation of LLMs has emerged as a critical area of research, with various studies highlighting the challenges and methodologies in assessing their performance. 
Reliable model evaluation is a critical problem. Traditional human evaluations remain the gold standard, but their scalability is a significant bottleneck in large-scale applications. Thus, recent works have proposed leveraging LLMs to evaluate the text quality, ranking outputs, and ensuring alignment with human preferences \citep{zheng2023judging, liu-etal-2023-g, dubois2024lengthcontrolledalpacaevalsimpleway}. While initially focused on text generation evaluation, the use of LLMs as judges has expanded to diverse applications including model alignment and safety assessment \citep{lee2024rlaifvsrlhfscaling}, code quality evaluation \citep{zhao2024codejudgeevallargelanguagemodels}, and knowledge verification \citep{min-etal-2023-factscore}, etc.
% , and multi-modal content assessment \citep{chen2024mllmasajudgeassessingmultimodalllmasajudge}. 
% \vspace{-3mm}

\begin{figure}[!t]
\includegraphics[width=1\columnwidth]{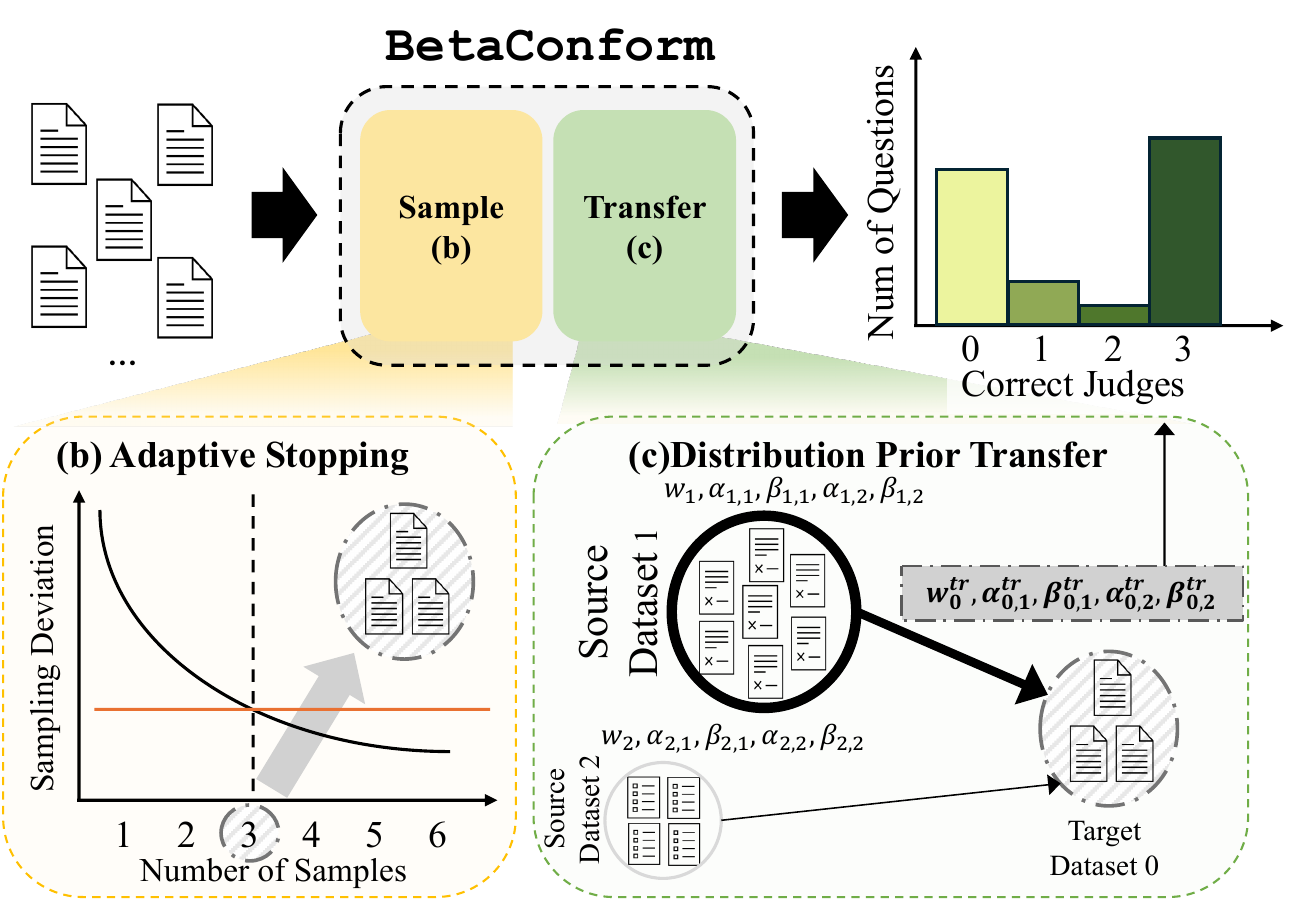}
\vspace{-7mm}
\caption{Overview of~\bc. Given a target dataset, adaptive stopping is adopted to determine the sample amount (\textbf{b}, Section \ref{sec:adaptive_stopping}). During iterative sampling, the sampling deviation is monitored by using conformal prediction. The sampling process stops when the deviation is sufficiently low. Next, the estimation of the small number of samples from the previous step is further enhanced by transferring distribution priors from source datasets (\textbf{c}, Section \ref{sec:transfer}). The transfer mechanism will assign a larger weight to the dataset that is textually closer to the target dataset.}
\label{fig:main_figure}
\vspace{-5mm}
\end{figure}

\paragraph{Challenges and Limitations.} The reliability of such frameworks is not without concerns. Studies have found that even advanced models like GPT-4 often exhibit systematic biases such as position bias and egocentric bias \citep{zeng2023evaluating, wang2023largelanguagemodelsfair}, overconfidence in their judgments \citep{koo2024benchmarkingcognitivebiaseslarge}, and self-preference effects \citep{panickssery2024llmevaluatorsrecognizefavor}. Moreover, many studies employing LLM annotations do not explicitly measure the alignment between LLMs and humans, thus further raising questions about their dependability \citep{calderon2025alternativeannotatortestllmasajudge}. While researchers have proposed various solutions, including dynamic evaluation pipelines \citep{yu2024kievalknowledgegroundedinteractiveevaluation, zhao2024autoarenaautomatingllmevaluations, moniri2024evaluatingperformancelargelanguage}, self-reflection mechanisms \citep{wu2024metarewardinglanguagemodelsselfimproving, li2023rainlanguagemodelsalign, wang2024selftaughtevaluators}, and specialized benchmarks for assessing judge performance \citep{zheng2023judging, tan2024judgebench, park2024offsetbiasleveragingdebiaseddata, li2024vlrewardbenchchallengingbenchmarkvisionlanguage, zhao2024codejudgeevallargelanguagemodels,li2024generation}, these methods often fall short in offering rigorous guarantees of their outcomes.
% \vspace{-3mm}
\paragraph{Statistical Approaches.} Another direction of research focuses on providing statistical guarantees for LLM performance. Researchers have explored conformal methods \citep{angelopoulos2023conformalriskcontrol} to ensure correctness and factuality \citep{mohri2024languagemodelsconformalfactuality} and to determine when LLMs should abstain from responding \citep{yadkori2024mitigatingllmhallucinationsconformal}. While these methods provide some statistical rigor, there is still a need for a unified framework that establishes reliable, theoretically grounded approaches for assessing LLM performance across diverse applications.

\section{Problem Setup}
\label{sec:setup}

We consider the task of using an LLM ensemble to evaluate and judge samples by discerning, choosing, or scoring. Let:
\begin{itemize}[noitemsep, topsep=0pt]
    \item \( n \): Total number of samples in the dataset.
    \item \( k \): Number of LLMs in an ensemble.
    \item \( S \): The random variable of correct judgments.
    \item \( r \): Number of samples to estimate $S$.
    \item \( D \): A dataset to estimate the judgment distribution.
\end{itemize}

\begin{definition}[LLM Ensemble Judgment]
Let $\mathcal{J} = \{J_1, J_2, \ldots, J_k\}$ be an ensemble of $k$ LLM judges. For a given input $x$, each LLM $J_i$ generates an output $o_i=J_i(x)$, producing all judgments as $\mathcal{O} = \{\,o_1, o_2, \ldots, o_k\}$. In this paper, we focus on binary and scoring judgment.
\end{definition}

% \begin{definition}[Correct judgment]
% Let $o_i$ be the judgment from the judge $J_i$ and $y$ be the answer, $o_i$ is correct if
% \begin{equation}
%     \mathrm{Match}(o_i,y)=1.
% \end{equation}
% The $\mathrm{Match}(\cdot)$ function can be exact matching, keyword matching or the distance to the groundtruth score, and it outputs $1$ when the matching criteria is met otherwise $0$.
% \end{definition}

\begin{definition}[LLM Ensemble Correct Judgment]
For an ensemble of $k$ LLMs, the random variable
\begin{equation}
    S=\sum_{i=1}^k\mathrm{Match}(o_i, y)
\end{equation}
denotes the number of correct judgments, where $y$ is the ground truth and $\mathrm{Match}(\cdot)$ is the criteria for correct judgment. The ensemble's decision is correct if $S \geq \lceil k / 2 \rceil$. \emph{To avoid the situation of a tie when $k$ is an even number and $S=\lceil k/2\rceil$, we only consider an odd number of $k$.}
\end{definition}

% \vspace{-3mm}
\section{Mixture of Beta-Binomial Distribution}

\subsection{Examination of Binomial Distribution}
We start by examining the common assumption of $S$ follows a Binomial distribution, i.e. the probability of having $s$ correct judgments when a single judge accuracy $\hat{p}$ is,
\begin{equation}
    \label{eq:binomial}
    \mathbb{P}_\mathrm{Bin}(S=s)=\mathrm{Bin}(s\mid k,\hat{p})= \binom{k}{s} \hat{p}^s (1-\hat{p})^{k-s}.
\end{equation}
The error rate $\Tilde{P}_{Bin}$ of ensemble judgment is:
\begin{equation}
    \label{eq:bin_error}
    \Tilde{P}_\mathrm{Bin} = \mathbb{P}_\mathrm{Bin}(S < \lceil k / 2 \rceil) = \sum_{s=0}^{\lceil k / 2 \rceil - 1} \binom{k}{s} \hat{p}^s (1-\hat{p})^{k-s}.
\end{equation}
We first examine the common assumption that $S$ follows a Binomial distribution in Equation (\ref{eq:binomial}). Specifically, we \ding{182} evaluate individual LLMs on datasets across domains and \ding{183} use the single LLM accuracy $p$ in Equation (\ref{eq:binomial}) and (\ref{eq:bin_error}) to estimate both the distribution of LLM ensembles on these datasets and the majority voting error rate for different numbers $k$ of LLMs. Specifically, we evaluate GPT-4 \cite{openai2024gpt4technicalreport} and Llama-3.3-70B \cite{dubey2024llama} on hallucination detection (HaluEval, \citealp{li2023halueval}) and Human alignment (JudgeBench, \citealp{tan2024judgebench}) datasets. Results are shown in Figure \ref{fig:dist_count} and Figure \ref{fig:dist_error}.

The results in Figure \ref{fig:dist_count} and Figure \ref{fig:dist_error} demonstrate the large deviation of Binomial distribution to the real distribution. On both datasets, the real distributions of LLM ensemble judgments consistently show two peaks centering at the two ends, while Binomial distribution results in a single peak with a large shift to either of the two peaks. Notably, in Figure \ref{fig:dist_error}, the assumption of a Binomial distribution leads to an always decreasing majority voting error rate, which is in sharp contrast with the actual error rate that remains at the same level when the ensemble becomes larger.

\subsection{Mixture of Beta-Binomial Distributions}
\begin{assumption}[Mixture of Beta-Binomial Distribuitons]
\label{eq:bb_dist}
\begin{equation}
    S \sim w\mathrm{BB}(k, \alpha_1, \beta_1) + (1-w)\mathrm{BB}(k, \alpha_2, \beta_2),
\end{equation}
where $\text{BB}(\cdot, \cdot, \cdot)$ is the Beta-Binomial distribution, $k$ is the number of judges in the ensemble, $\alpha_1, \beta_1, \alpha_2, \beta_2$ are parameters of the two distributions, and $w$ is the mixture weight.
\end{assumption}

\begin{corollary}[Mixture Distribution Error Rate]
The error rate of the mixture of Beta-Binomial distributions is
\begin{equation}
    \label{eq:BB_error}
    \begin{aligned}
    \Tilde{P}_\mathrm{BB}=w \sum_{s=0}^{\lceil k / 2 \rceil - 1}\binom{k}{s}\frac{\mathrm{B}(s+\alpha_1,\;k-s+\beta_1)}{\mathrm{B}(\alpha_1,\beta_1)}\\
    +(1-w) \sum_{s=0}^{\lceil k / 2 \rceil - 1}\binom{k}{s}\frac{\mathrm{B}(s+\alpha_2,\;k-s+\beta_2)}{\mathrm{B}(\alpha_2,\beta_2)},
    \end{aligned}
\end{equation}
where $\mathrm{B}(\cdot, \cdot)$ is the Beta function.
\end{corollary}

\begin{figure*}[t]
    \centering
    \includegraphics[width=1\linewidth]{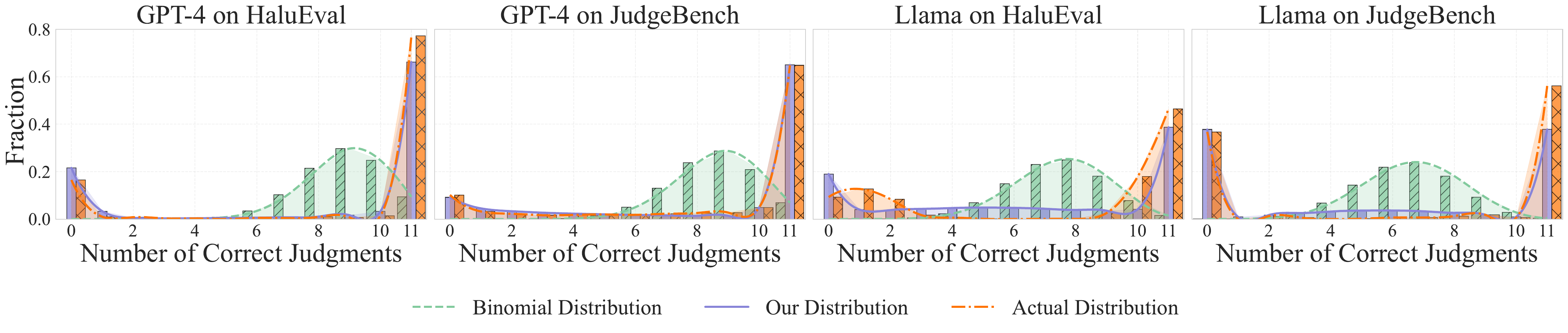}
    \vspace{-8mm}
    \caption{Comparison of judgment distributions among actual, Binomial, and ours. Llama-3.3-70B and GPT-4 ensembles of $11$ models are tested on HaluEval and JudgeBench, respectively. The Binomial distribution is estimated by using single judge accuracy $p$. Our mixture distribution is estimated with $100$ samples and scaled to the full dataset. \textbf{Our distribution is consistently closer to the actual one.}}
    \label{fig:dist_count}
\end{figure*}

\begin{figure*}[t]
    \centering
    \includegraphics[width=1\linewidth]{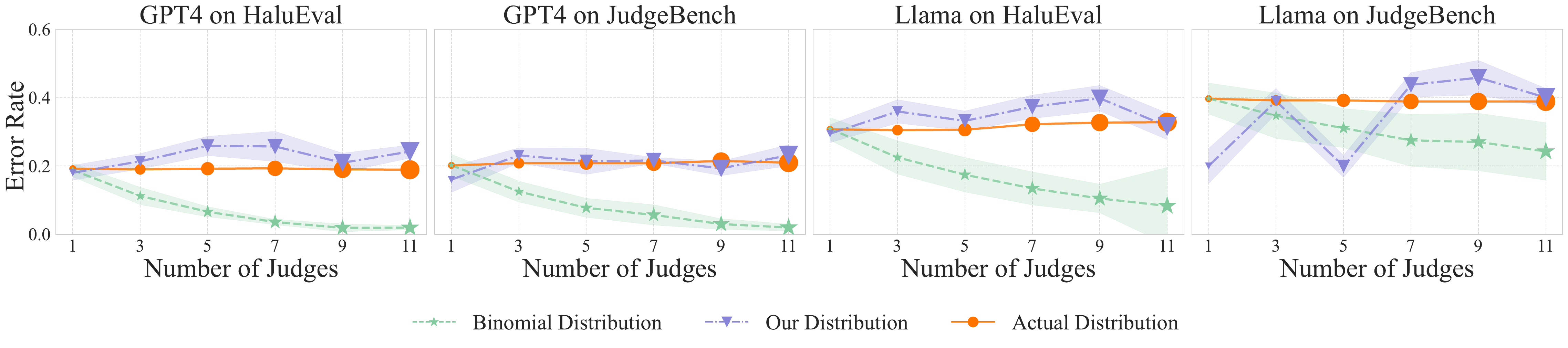}
    \vspace{-8mm}
    \caption{Majority voting error rate of actual, Binomial, and our mixture distribution. Binomial uses single judge accuracy $p$. Our distribution is estimated with $100$ random samples and tested for $3$ times. The line denotes the average error rate and the shadow represents the standard variance. Binomial shows decreasing error rate, while \textbf{our distribution captures the actual trend.}}
    \label{fig:dist_error}
\end{figure*}

After examining the common Binomial distribution assumption in Figure \ref{fig:dist_count} and Figure \ref{fig:dist_error}, we notice that the real distribution keeps showing two peaks centering near all wrong and all correct. Motivated by this observation, in Assumption \ref{eq:bb_dist} we model the distribution as a mixture of two Beta-Binomial distributions, where one distribution models the LLM ensemble judgments on simple questions and the other one for hard problems. To derive all the parameters, we utilize labeled samples from the dataset and design a distribution-tailored expectation maximization (EM) algorithm.

\subsection{Expectation Maximization}
\label{app:em}
\paragraph{Samples as Distribution Evidence.}
Given $r$ samples, each containing judgments from $k$ LLMs, $S_i$ is the number of correct judgments in the $i$-th sample and $p_i=S_i/k$ as the estimated probability of success for the $i$-th sample.

% The estimation of the mixture distribution starts from estimating the probability distribution of judging with a single LLM. Under Assumption \ref{eq:bb_dist}, the single LLM distribution is a mixture of two Beta distributions. In the $r$ samples, each sample contributes probabilistically to both Beta distributions, weighted by their likelihood under each component.

For the $i$-th sample, considering the first Beta-Binomial distribution, a responsibility $\gamma_1^i$ is assigned to it

\vspace{-5mm}
\begin{equation}
    \gamma_1^i = \frac{w \mathrm{Beta}(p_i \mid \alpha_1, \beta_1)}{w \mathrm{Beta}(p_i \mid \alpha_1, \beta_1) + (1-w) \mathrm{Beta}(p_i \mid \alpha_2, \beta_2)},
\end{equation}
where $\mathrm{Beta}(p_i \mid \alpha, \beta)$ is the probability density of Beta distribution at $p_i$ for the $i$-th sample under the corresponding Beta component.
$\gamma_1^i$ represents the probability that the $i$-th sample belongs to the first Beta component, and $\gamma_2^i = 1 - \gamma_1^i$ is the probability for the second component.

\paragraph{Parameters Update.}
The parameters are updated based on the weighted contributions of samples. The parameters of two distributions $j=\{1, 2\}$ are updated as
\begin{equation}
\begin{gathered}
\alpha_j^\prime = \sum_{i=1}^r \gamma_1^i \cdot S_i,\ \beta_j^\prime = \sum_{i=1}^r \gamma_1^i \cdot (k-S_i),\ w^\prime = \frac{1}{r} \sum_{i=1}^r \gamma_1^i
\end{gathered}
\end{equation}
% The weight $w$ in 

We verify our distribution assumption by first sampling $r=100$ judgments made by two models on two datasets and apply our distribution-tailored EM algorithm to estimate the parameters. Our method is evaluated in two scenarios: \ding{182} In Figure \ref{fig:dist_count}, we fix the ensemble size $k=11$ and compare the estimated distribution against the real distribution and Binomial distribution, and \ding{183} in Figure \ref{fig:dist_error} we estimate the error rate of majority voting with different ensemble sizes.

In Figure \ref{fig:dist_count}, the mixture of Beta-Binomial distributions is significantly closer to the real distribution compared to the Binomial, with clear two-peak patterns that are analogous to the observation. In Figure \ref{fig:dist_error} it shows that our distribution is consistently close to the real majority voting error rate across all ensemble sizes. Contrary to the Binomial distribution that produced a decreasing error rate, our distribution successfully modeled the stable error rate when the ensemble becomes larger. Additionally, the narrow confidence interval demonstrates the high stability of our method.

\section{Guide Sampling via Conformal Prediction}
\label{sec:adaptive_stopping}
In the experiments above, we used a fixed number of samples. However, in practical settings where datasets are unannotated and being labeled, it is essential to determine when the number of annotated samples is sufficient for accurate estimation. Inspired by conformal prediction (CP), which does not rely on prior knowledge of the dataset distribution and can rigorously estimate the sampling deviation, we propose leveraging its principles to address this challenge.

\subsection{Conformal Prediction for Adaptive Stopping}
CP provides a principled approach to dynamically evaluate the sampling deviation in the distribution of the number of correct judgments $S$, which can be used as guidance.

\vspace{-5mm}
\paragraph{Nonconformity Scores.} A major part of CP is the nonconformity score, which measures how a test sample differs from the rest of the data. In our implementation, we set the nonconformity score as 
\begin{equation}
    \label{eq:nonconformity_score}
    \mathrm{score}(S_i) = |S_i - \mathbb{E}[S]|,
\end{equation}
which quantifies the deviation of each observed value of $S$ from the expected value.

\paragraph{Calibration Data and Quantile Computation.}
Suppose $r$ samples have been used to test the LLM ensemble with $S_1, S_2, \ldots, S_r$ correct judgments, the CP sampling computes the nonconformity scores for all calibration data as $s_i=\mathrm{score}(S_{i})$ and these scores are sorted in ascending order as $s_1<\ldots<s_r$. For a desired estimation confidence \( 1 - \epsilon \), the \( (1 - \epsilon) \)-quantile with $r$ samples \( q_{1-\epsilon}^r \) is
\begin{equation}
    q_{1-\epsilon}^r = s_{\lceil (1 - \epsilon) \cdot (r + 1) \rceil}.
    \vspace{-3mm}
\end{equation}

\begin{figure*}[h]
    \centering
    \includegraphics[width=1\linewidth]{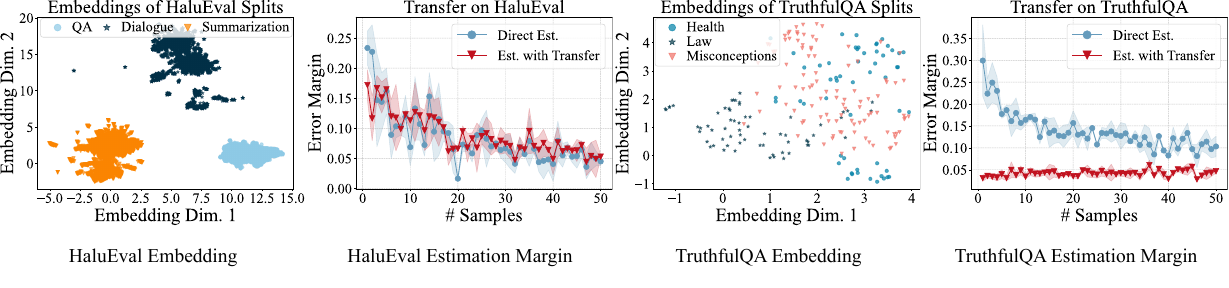}
    \vspace{-8mm}
    \caption{Examples of distribution prior transfer. Splits from HaluEval form distinct clusters in the embedding space, and transfer does not degrade performance compared to only using target dataset samples. In contrast, topics in TruthfulQA exhibit closer proximity, where transfer leads to significant performance improvements compared to solely using the limited samples of the target dataset.}
    \label{fig:transfer}
    \vspace{-5mm}
\end{figure*}

\paragraph{Adaptive Stopping Criteria.}
Adaptive stopping is achieved by monitoring the variation of the conformal prediction quantile. After \( r \) samples, the $(1-\epsilon)$-quantile is recomputed and compared with the one from $r-1$ samples. The sampling process stops when the quantile satisfies
\begin{equation}
    \label{eq:stop_criteria}
    \left|q_{1-\epsilon}^r-q_{1-\epsilon}^{r-1}\right| \leq \xi
\end{equation}
where \( \xi \) is a predefined threshold.

\begin{proposition}[Sample Amount with Adaptive Stopping]
\label{pro:amount}
For a given sampling deviation threshold $\xi$ and a scale $\tau$, the sample amount $r$ should satisfy 
\begin{equation}
    \label{eq:sample_number}
    \tau\left(\frac{1}{\sqrt{r-1}}-\frac{1}{\sqrt{r}}\right) \leq \xi,
\end{equation}

\end{proposition}
This proposition offers an estimation of the sample amount under the threshold $\xi$.

\begin{proposition}[Error Rate with Adaptive Stopping]
\label{pro:error_rate}
Under the sampling threshold $\xi$, the majority voting error rate of the mixture distribution becomes
\begin{equation}
    \label{eq:stopping_error}
     (1-\min(\xi, \frac{\tau}{\sqrt{r}}))\Tilde{P}_\mathrm{BB} < \Tilde{P}_{adapt} < (1+\min(\xi, \frac{\tau}{\sqrt{r}}))\Tilde{P}_\mathrm{BB}
    % \Tilde{P}_{adpt} = (1 \pm \xi)\Tilde{P}_\mathrm{BB}
\end{equation}
\end{proposition}
This proposition provides a theoretical error bound for estimation under adaptive stopping, suggesting the mild degradation of estimation performance.
% \begin{corollary}[Error Bound under Adaptive Stopping]
% Given Equation (\ref{eq:stopping_error}), the estimation error between using the full dataset and samples is bounded as
% \begin{equation}
% |\Tilde{P}_{adapt} - \Tilde{P}_\mathrm{BB} |\leq \min(\xi, \frac{\tau}{\sqrt{r}})\Tilde{P}_{adpt}
% % |\Tilde{P}_{adpt} - \Tilde{P}_\mathrm{BB}| \leq \xi \Tilde{P}_{adpt}
% \end{equation}
% \end{corollary}

We leave the proofs of Proposition \ref{pro:amount} and \ref{pro:error_rate} in Appendix \ref{app:stopping} and \ref{app: error}, respectively. In our experiments, we set $\xi=0.03$, and $\tau=25$, which leads to $r\geq56$.

\vspace{-1em}
\section{Text Similarity for Distribution Prior Transfer}
\label{sec:transfer}
To further improve the data efficiency when only a few samples are available and enhance estimation accuracy, we propose to incorporate prior knowledge about the LLM ensemble on other open-source datasets and transfer the estimated judgment distributions to the target dataset. However, one challenge is that the prior transfer could bring performance degradation if the distributions of the source datasets and the target dataset are very different. To resolve this challenge, we design text similarity-based distribution prior transfer, which leverages the strong text embedding capability of the recent models to understand and measure the textual difference among datasets.

\vspace{-1em}
\paragraph{Text Embedding.} To embed the text inputs of the LLM ensemble, we use NV-Embed-V2 \cite{lee2025nvembedimprovedtechniquestraining}. Given sets of samples $\left\{D_1, D_2, \ldots, D_m\right\}$ from $m$ source datasets, the embedding model $\mathcal{E}(\cdot)$ is utilized to transform the sets of samples to sets of embeddings for the source datasets
\begin{equation}
    \left\{E_1, E_2, \ldots, E_m\right\} = \left\{\mathcal{E}(D_1),\mathcal{E}(D_2),\ldots,\mathcal{E}(D_m)\right\}.
\end{equation}
The average embedding $\bar{E_i}=\frac{1}{r_i}\sum_{j=1}^{r_i}E_i^j$ of the $i$-th dataset is used to represent it.

\vspace{-1em}
\paragraph{Distribution Prior Transfer.} To transfer the distribution from source datasets to the target dataset $D_0$, the process starts by embedding the target dataset $E_0=\mathcal{E}(D_0)$ and acquiring its average embedding $\bar{E_0}$. For the dataset $D_i$, its transfer weight is
\begin{equation}
    \label{eq:tr_weight}
    \lambda_i=\log(r_i)\cdot\sigma\left(\alpha\cdot\left(\mathrm{CosSim}\left(\bar{E_0},\bar{E_i}\right)-\beta\right)\right),
\end{equation}
where $\sigma(\cdot)$ is the sigmoid function, $r_i$ is the number of samples and $\alpha$ and $\beta$ are hyperparameters. We adopt this design to avoid the degradation of estimation caused by transferring datasets with dissimilar text inputs. This is achieved by setting a threshold and applying the sigmoid function to suppress the weight when the similarity is low. $\log(r_i)$ is included as datasets with more samples could produce a more accurate estimation and thus should have a higher impact on the transfer. The transfer is performed as
\begin{equation}
\label{eq:transfer}
\begin{aligned}
    w^{tr}_0 &= \frac{\sum_{i=0}^m \lambda_i \cdot w_i}{\sum_{i=0}^m \lambda_i},
    \alpha_{0,j}^{tr} = \frac{\sum_{i=0}^m \lambda_i \cdot \alpha_{i,j}}{\sum_{i=0}^m \lambda_i}, \\
    \beta_{0,j}^{tr} &= \frac{\sum_{i=0}^m \lambda_i \cdot \beta_{i,j}}{\sum_{i=0}^m \lambda_i},
    \quad j \in \{1,2\}.
\end{aligned}
\end{equation}

In Equation (\ref{eq:transfer}), $\alpha_{i,j}$ and $\beta_{i,j}$ are the $j$-th parameter in the mixture distribution of $i$-th dataset. The parameters in the weighted sum with index $0$ denote direct estimation using the limited samples of the target dataset.
\paragraph{Examples.} To verify our distribution design, we evaluate the distribution within splits of HaluEval \cite{li2023halueval} and TruthfulQA \cite{lin2021truthfulqa} datasets. For HaluEval, we use Dialogue and Summarization splits as source datasets and transfer to QA split; for TruthfulQA, we transfer from topics of Health and Law to Misconceptions. As shown in Figure \ref{fig:transfer}, the embeddings form distant clusters in HaluEval, as the text inputs of the three splits have different hallucination detection requirements, and embeddings from TruthfulQA overlap due to the similarity of judgment format. When clusters are separated, our method will not bring performance degradation compared to solely using samples from the target dataset, while when clusters are overlapping, our method brings a significantly lower estimation error rate margin compared to only using target dataset samples. This supports the effectiveness of our distribution transfer design.

\vspace{-1em}
\section{\bc}
In this section, we present \bc, the framework for efficient estimation of judgment distribution, as shown in Figure \ref{fig:main_figure} and Algorithm \ref{alg:bc}. When only limited samples on the target dataset are available, it transfers distribution priors from source datasets. Given more samples, it employs adaptive stopping during iterative sampling to balance sample efficiency with estimation accuracy.

\vspace{-3mm}
\begin{algorithm}
   \caption{\bc}
   \label{alg:bc}
\begin{algorithmic}[1]
   \STATE {\bfseries Input:} \parbox[t]{\textwidth}{target dataset $D_0$, source datasets $D_1, \ldots,D_m$,\\ judges $\mathcal{J}=\{J_1, \ldots, J_k\}$, EM algorithm $\mathrm{EM(\cdot)}$}
   \STATE {\bfseries Output:} distribution parameters $\Omega$ on the target dataset
   \IF{limited samples in $D_0$}
   \STATE Compute distribution parameters on $D_0$
   \STATE Compute parameters of distributions on $D_1, \ldots,D_m$
   \STATE Compute transfer weights by Equation (\ref{eq:tr_weight})
   \STATE $\Omega \leftarrow$ Compute transferred parameters by Eq. (\ref{eq:transfer})
   \ELSE
   \STATE Initial $D \leftarrow \{\}$, $q_{1-\epsilon}^0 \leftarrow -\infty$
   \WHILE{Equation (\ref{eq:stop_criteria}) is not satisfied}
   \STATE Add a sample from $D_0$ to $D$ and update $q_{1-\epsilon}^{\left|D\right|}$
   \ENDWHILE
   \STATE $\Omega \leftarrow$ Compute distribution parameters on samples $D$
   \ENDIF
   \STATE {\bfseries return} $\Omega$
\end{algorithmic}
\end{algorithm}

\vspace{-5mm}
\section{Experiments}
We evaluate LLM ensembles of up to $11$ models, including popular \textit{close-source} models (GPT-3.5, \citealp{brown2020language}; GPT-4, \citealp{openai2024gpt4technicalreport}) and \textit{open-source} models (Llama-3.3-70B, \citealp{dubey2024llama}; Qwen-2.5-72B, \citealp{yang2024qwen2}; InternLM-2.5-20B, \citealp{cai2024internlm2}). We choose domains of \textit{hallucination detection} (HaluEval, \citealp{li2023halueval}; TruthfulQA, \citealp{lin2021truthfulqa}; HalluDial, \citealp{luo2024halludial}), \textit{reasoning} (PRM800K, \citealp{lightman2023let}; BIG-bench, \citealp{srivastava2022beyond}; TRAM, \citealp{wang2023tram}), \textit{scoring} (ICE-Score, \citealp{zhuo2023ice}; Comp-Analysis, \citealp{zhang2024comprehensive}) and \textit{alignment} (JudgeBench, \citealp{tan2024judgebench}; RewardBench, \citealp{lambert2024rewardbench}; LLMBar, \citealp{zeng2023evaluating}). Across experiments, we set the adaptive stopping threshold to $\xi=0.01$, which requires at least $r\geq51$ samples to meet the stopping criteria. The sampling and estimation process is conducted for $30$ times on each dataset.

\begin{table*}[!t]
\caption{The comparison of error margins between our mixture of Beta-Binomial distributions and Binomial distribution. The \textbf{Err. Margin} and \textbf{\# Samples} answer \textbf{\hyperref[rq1]{RQ1}} and \textbf{\hyperref[rq2]{RQ2}}, respectively. The error margin is calculated as the absolute difference between the actual error rate and the estimation. Estimations using both distributions are done on samples obtained through iterative sampling with adaptive stopping. For each run, the error margin is computed from $k=1$ to $11$, and the average margin of ensemble sizes is used as the result for that run. We conduct $30$ runs and report the average and standard deviation. The average number of samples across runs is also reported.}
\label{tab:est}
\centering
\resizebox{0.99\linewidth}{!}{%
\begin{tabular}{@{}llcccccccccc@{}}
\toprule
\multicolumn{1}{l|}{} &
  \multicolumn{1}{l|}{} &
  \multicolumn{2}{c|}{Llama-3.3-70B} &
  \multicolumn{2}{c|}{Qwen-2.5-72B} &
  \multicolumn{2}{c|}{InternLM-2.5-20B} &
  \multicolumn{2}{c|}{GPT-3.5} &
  \multicolumn{2}{c}{GPT-4} \\ \cmidrule(l){3-12} 
\multicolumn{1}{l|}{\multirow{-2}{*}{Dataset}} &
  \multicolumn{1}{l|}{\multirow{-2}{*}{Method}} &
  Err. Margin~($\downarrow$) &
  \multicolumn{1}{c|}{\# Samples~($\downarrow$)} &
  Err. Margin~($\downarrow$) &
  \multicolumn{1}{c|}{\# Samples~($\downarrow$)} &
  Err. Margin~($\downarrow$) &
  \multicolumn{1}{c|}{\# Samples~($\downarrow$)} &
  Err. Margin~($\downarrow$) &
  \multicolumn{1}{c|}{\# Samples~($\downarrow$)} &
  Err. Margin~($\downarrow$) &
  \# Samples~($\downarrow$) \\ \midrule
\multicolumn{12}{c}{\cellcolor[HTML]{F5EBE0}Hallucination Detection Datasets} \\ \midrule
\multicolumn{1}{l|}{} &
  \multicolumn{1}{l|}{Binomial} &
  17.62 ± 0.73 &
  \multicolumn{1}{c|}{} &
  12.45 ± 1.04 &
  \multicolumn{1}{c|}{} &
  16.67 ± 0.38 &
  \multicolumn{1}{c|}{} &
  5.78 ± 0.08
   & 
  \multicolumn{1}{c|}{} &
  9.16 ± 0.18
   & 
   \\
\multicolumn{1}{l|}{\multirow{-2}{*}{HaluEval}} &
  \multicolumn{1}{l|}{\cellcolor[HTML]{EDEDE9}Ours} &
  \cellcolor[HTML]{EDEDE9}\textbf{6.68 ± 0.53} &
  \multicolumn{1}{c|}{\multirow{-2}{*}{49.47}} &
  \cellcolor[HTML]{EDEDE9}\textbf{4.72 ± 0.38} &
  \multicolumn{1}{c|}{\multirow{-2}{*}{61.02}} &
  \cellcolor[HTML]{EDEDE9}\textbf{5.48 ± 0.41} &
  \multicolumn{1}{c|}{\multirow{-2}{*}{50.67}} &
  \cellcolor[HTML]{EDEDE9} \textbf{5.10 ± 0.24}
  &
  \multicolumn{1}{c|}{\multirow{-2}{*}{34.80}} &
  \cellcolor[HTML]{EDEDE9} \textbf{6.28 ± 0.39} &
  \multirow{-2}{*}{40.58} \\ \midrule
\multicolumn{1}{l|}{} &
  \multicolumn{1}{l|}{Binomial} &
  14.00 ± 0.65 &
  \multicolumn{1}{c|}{} &
  19.86 ± 0.40 &
  \multicolumn{1}{c|}{} &
  19.55 ± 0.65 &
  \multicolumn{1}{c|}{} &
    14.44 ± 0.40   & 
  \multicolumn{1}{c|}{} &
  15.20 ± 0.55
   & 
   \\
\multicolumn{1}{l|}{\multirow{-2}{*}{TruthfulQA}} &
  \multicolumn{1}{l|}{\cellcolor[HTML]{EDEDE9}Ours} &
  \cellcolor[HTML]{EDEDE9}\textbf{7.53 ± 0.55} &
  \multicolumn{1}{c|}{\multirow{-2}{*}{54.13}} &
  \cellcolor[HTML]{EDEDE9}\textbf{7.18 ± 0.44} &
  \multicolumn{1}{c|}{\multirow{-2}{*}{53.56}} &
  \cellcolor[HTML]{EDEDE9}\textbf{6.24 ± 0.59} &
  \multicolumn{1}{c|}{\multirow{-2}{*}{55.56}} &
  \cellcolor[HTML]{EDEDE9}\textbf{6.75 ± 0.58} &
  \multicolumn{1}{c|}{\multirow{-2}{*}{47.64}} &
  \cellcolor[HTML]{EDEDE9}\textbf{6.73 ± 0.38} &
  \multirow{-2}{*}{57.07} \\ \midrule
\multicolumn{1}{l|}{} &
  \multicolumn{1}{l|}{Binomial} &
  13.10 ± 0.37 &
  \multicolumn{1}{c|}{} &
  13.42 ± 0.54 &
  \multicolumn{1}{c|}{} &
  14.84 ± 0.42 &
  \multicolumn{1}{c|}{} &
  8.79 ± 0.21
   &
  \multicolumn{1}{c|}{} &
  9.25 ± 0.27
   &
   \\
\multicolumn{1}{l|}{\multirow{-2}{*}{HalluDial}} &
  \multicolumn{1}{l|}{\cellcolor[HTML]{EDEDE9}Ours} &
  \cellcolor[HTML]{EDEDE9}\textbf{7.94 ± 0.68} &
  \multicolumn{1}{c|}{\multirow{-2}{*}{46.58}} &
  \cellcolor[HTML]{EDEDE9}\textbf{6.96 ± 0.47} &
  \multicolumn{1}{c|}{\multirow{-2}{*}{55.78}} &
  \cellcolor[HTML]{EDEDE9}\textbf{6.43 ± 0.50} &
  \multicolumn{1}{c|}{\multirow{-2}{*}{51.87}} &
  \cellcolor[HTML]{EDEDE9}\textbf{6.27 ± 0.36} &
  \multicolumn{1}{c|}{\multirow{-2}{*}{41.51}} &
  \cellcolor[HTML]{EDEDE9}\textbf{5.22 ± 0.59} &
  \multirow{-2}{*}{42.31} \\ \midrule
\multicolumn{12}{c}{\cellcolor[HTML]{F5EBE0}Reasoning Datasets} \\ \midrule
\multicolumn{1}{l|}{} &
  \multicolumn{1}{l|}{Binomial} &
  10.11 ± 0.29 &
  \multicolumn{1}{c|}{} &
  9.14 ± 0.17 &
  \multicolumn{1}{c|}{} &
  9.12 ± 0.20 &
  \multicolumn{1}{c|}{} &
  8.83 ± 0.25
   &
  \multicolumn{1}{c|}{} &
  14.52 ± 0.73
   &
   \\
\multicolumn{1}{l|}{\multirow{-2}{*}{PRM800K}} &
  \multicolumn{1}{l|}{\cellcolor[HTML]{EDEDE9}Ours} &
  \cellcolor[HTML]{EDEDE9}\textbf{9.37 ± 0.64} &
  \multicolumn{1}{c|}{\multirow{-2}{*}{43.33}} &
  \cellcolor[HTML]{EDEDE9}\textbf{7.82 ± 0.69} &
  \multicolumn{1}{c|}{\multirow{-2}{*}{42.89}} &
  \cellcolor[HTML]{EDEDE9}\textbf{4.52 ± 0.50} &
  \multicolumn{1}{c|}{\multirow{-2}{*}{46.13}} &
  \cellcolor[HTML]{EDEDE9} \textbf{8.46 ± 0.51}&
  \multicolumn{1}{c|}{\multirow{-2}{*}{51.38}} &
  \cellcolor[HTML]{EDEDE9}\textbf{6.17 ± 0.48} &
  \multirow{-2}{*}{54.67} \\ \midrule
\multicolumn{1}{l|}{} &
  \multicolumn{1}{l|}{Binomial} &
  13.29 ± 0.78 &
  \multicolumn{1}{c|}{} &
  14.17 ± 0.40 &
  \multicolumn{1}{c|}{} &
  14.68 ± 0.24 &
  \multicolumn{1}{c|}{} &
  14.83 ± 0.53
   &
  \multicolumn{1}{c|}{} &
  12.15 ± 0.74
   &
   \\
\multicolumn{1}{l|}{\multirow{-2}{*}{BIG-bench}} &
  \multicolumn{1}{l|}{\cellcolor[HTML]{EDEDE9}Ours} &
  \cellcolor[HTML]{EDEDE9}\textbf{11.15 ± 0.60} &
  \multicolumn{1}{c|}{\multirow{-2}{*}{51.51}} &
  \cellcolor[HTML]{EDEDE9}\textbf{6.97 ± 0.58} &
  \multicolumn{1}{c|}{\multirow{-2}{*}{47.82}} &
  \cellcolor[HTML]{EDEDE9}\textbf{5.54 ± 0.51} &
  \multicolumn{1}{c|}{\multirow{-2}{*}{48.40}} &
  \cellcolor[HTML]{EDEDE9}\textbf{12.59 ± 0.48} &
  \multicolumn{1}{c|}{\multirow{-2}{*}{46.13}} &
  \cellcolor[HTML]{EDEDE9}\textbf{8.02 ± 0.59} &
  \multirow{-2}{*}{46.09} \\ \midrule
\multicolumn{1}{l|}{} &
  \multicolumn{1}{l|}{Binomial} &
  14.79 ± 0.82 &
  \multicolumn{1}{c|}{} &
  13.13 ± 0.64 &
  \multicolumn{1}{c|}{} &
  13.06 ± 0.77 &
  \multicolumn{1}{c|}{} &
  4.99 ± 0.13
   &
  \multicolumn{1}{c|}{} &
  5.14 ± 0.11
   &
   \\
\multicolumn{1}{l|}{\multirow{-2}{*}{TRAM}} &
  \multicolumn{1}{l|}{\cellcolor[HTML]{EDEDE9}Ours} &
  \cellcolor[HTML]{EDEDE9}\textbf{8.39 ± 0.63} &
  \multicolumn{1}{c|}{\multirow{-2}{*}{55.87}} &
  \cellcolor[HTML]{EDEDE9}\textbf{6.20 ± 0.34} &
  \multicolumn{1}{c|}{\multirow{-2}{*}{57.16}} &
  \cellcolor[HTML]{EDEDE9}\textbf{6.10 ± 0.58} &
  \multicolumn{1}{c|}{\multirow{-2}{*}{57.78}} &
  \cellcolor[HTML]{EDEDE9}\textbf{3.94 ± 0.17} &
  \multicolumn{1}{c|}{\multirow{-2}{*}{39.07}} &
  \cellcolor[HTML]{EDEDE9}\textbf{4.81 ± 0.23} &
  \multirow{-2}{*}{38.53} \\ \midrule
\multicolumn{12}{c}{\cellcolor[HTML]{F5EBE0}Human Alignment Datasets} \\ \midrule
\multicolumn{1}{l|}{} &
  \multicolumn{1}{l|}{Binomial} &
  12.06 ± 0.78 &
  \multicolumn{1}{c|}{} &
  13.45 ± 0.54 &
  \multicolumn{1}{c|}{} &
  10.31 ± 1.03 &
  \multicolumn{1}{c|}{} &
  8.85 ± 0.33
   &
  \multicolumn{1}{c|}{} &
  10.98 ± 0.32
   &
   \\
\multicolumn{1}{l|}{\multirow{-2}{*}{JudgeBench}} &
  \multicolumn{1}{l|}{\cellcolor[HTML]{EDEDE9}Ours} &
  \cellcolor[HTML]{EDEDE9}\textbf{6.98 ± 0.56} &
  \multicolumn{1}{c|}{\multirow{-2}{*}{60.58}} &
  \cellcolor[HTML]{EDEDE9}\textbf{5.39 ± 0.39} &
  \multicolumn{1}{c|}{\multirow{-2}{*}{58.40}} &
  \cellcolor[HTML]{EDEDE9}\textbf{5.26 ± 0.39} &
  \multicolumn{1}{c|}{\multirow{-2}{*}{57.16}} &
  \cellcolor[HTML]{EDEDE9}\textbf{7.03 ± 0.61} &
  \multicolumn{1}{c|}{\multirow{-2}{*}{41.07}} &
  \cellcolor[HTML]{EDEDE9}\textbf{6.45 ± 0.53} &
  \multirow{-2}{*}{46.58} \\ \midrule
\multicolumn{1}{l|}{} &
  \multicolumn{1}{l|}{Binomial} &
  \textbf{8.40 ± 0.19} &
  \multicolumn{1}{c|}{} &
  8.93 ± 0.22 &
  \multicolumn{1}{c|}{} &
  17.36 ± 1.41 &
  \multicolumn{1}{c|}{} &
  11.42 ± 0.33
   &
  \multicolumn{1}{c|}{} &
  13.98 ± 0.29
   &
   \\
\multicolumn{1}{l|}{\multirow{-2}{*}{RewardBench}} &
  \multicolumn{1}{l|}{\cellcolor[HTML]{EDEDE9}Ours} &
  \cellcolor[HTML]{EDEDE9}11.30 ± 0.62 &
  \multicolumn{1}{c|}{\multirow{-2}{*}{40.22}} &
  \cellcolor[HTML]{EDEDE9}\textbf{4.68 ± 0.56} &
  \multicolumn{1}{c|}{\multirow{-2}{*}{45.20}} &
  \cellcolor[HTML]{EDEDE9}\textbf{6.58 ± 0.40} &
  \multicolumn{1}{c|}{\multirow{-2}{*}{52.04}} &
  \cellcolor[HTML]{EDEDE9}\textbf{6.90 ± 0.45} &
  \multicolumn{1}{c|}{\multirow{-2}{*}{42.27}} &
  \cellcolor[HTML]{EDEDE9}\textbf{7.65 ± 0.51} &
  \multirow{-2}{*}{48.22} \\ \midrule
\multicolumn{1}{l|}{} &
  \multicolumn{1}{l|}{Binomial} &
  13.61 ± 0.58 &
  \multicolumn{1}{c|}{} &
  14.63 ± 0.51 &
  \multicolumn{1}{c|}{} &
  13.66 ± 1.14 &
  \multicolumn{1}{c|}{} &
  \textbf{13.19 ± 0.55}
   &
  \multicolumn{1}{c|}{} &
  10.36 ± 0.33
   &
   \\
\multicolumn{1}{l|}{\multirow{-2}{*}{LLMBar}} &
  \multicolumn{1}{l|}{\cellcolor[HTML]{EDEDE9}Ours} &
  \cellcolor[HTML]{EDEDE9}\textbf{10.18 ± 0.71} &
  \multicolumn{1}{c|}{\multirow{-2}{*}{50.18}} &
  \cellcolor[HTML]{EDEDE9}\textbf{7.52 ± 0.63} &
  \multicolumn{1}{c|}{\multirow{-2}{*}{51.07}} &
  \cellcolor[HTML]{EDEDE9}\textbf{6.38 ± 0.53} &
  \multicolumn{1}{c|}{\multirow{-2}{*}{51.29}} &
  \cellcolor[HTML]{EDEDE9}13.71 ± 0.54 &
  \multicolumn{1}{c|}{\multirow{-2}{*}{44.40}} &
  \cellcolor[HTML]{EDEDE9}\textbf{8.16 ± 0.50} &
  \multirow{-2}{*}{44.40} \\ \midrule
\multicolumn{12}{c}{\cellcolor[HTML]{F5EBE0}Scoring Datasets} \\ \midrule
\multicolumn{1}{l|}{} &
  \multicolumn{1}{l|}{Binomial} &
  \textbf{8.91 ± 0.25} &
  \multicolumn{1}{c|}{} &
  9.27 ± 0.23 &
  \multicolumn{1}{c|}{} &
  22.24 ± 1.02 &
  \multicolumn{1}{c|}{} &
  3.61 ± 0.06
   &
  \multicolumn{1}{c|}{} &
  \textbf{3.66 ± 0.07}
   &
   \\
\multicolumn{1}{l|}{\multirow{-2}{*}{ICE-Score}} &
  \multicolumn{1}{l|}{\cellcolor[HTML]{EDEDE9}Ours} &
  \cellcolor[HTML]{EDEDE9}8.97 ± 0.45 &
  \multicolumn{1}{c|}{\multirow{-2}{*}{41.29}} &
  \cellcolor[HTML]{EDEDE9}\textbf{6.91 ± 0.59} &
  \multicolumn{1}{c|}{\multirow{-2}{*}{43.73}} &
  \cellcolor[HTML]{EDEDE9}\textbf{18.19 ± 0.37} &
  \multicolumn{1}{c|}{\multirow{-2}{*}{53.42}} &
  \cellcolor[HTML]{EDEDE9}\textbf{3.39 ± 0.32} &
  \multicolumn{1}{c|}{\multirow{-2}{*}{39.87}} &
  \cellcolor[HTML]{EDEDE9}5.78 ± 0.08 &
  \multirow{-2}{*}{38.93} \\ \midrule
\multicolumn{1}{l|}{} &
  \multicolumn{1}{l|}{Binomial} &
  14.45 ± 0.71 &
  \multicolumn{1}{c|}{} &
  15.88 ± 0.72 &
  \multicolumn{1}{c|}{} &
  13.28 ± 0.73 &
  \multicolumn{1}{c|}{} &
  12.87 ± 0.32
   &
  \multicolumn{1}{c|}{} &
  15.64 ± 0.68
   &
   \\
\multicolumn{1}{l|}{\multirow{-2}{*}{COMP-Analysis}} &
  \multicolumn{1}{l|}{\cellcolor[HTML]{EDEDE9}Ours} &
  \cellcolor[HTML]{EDEDE9}\textbf{6.50 ± 0.63} &
  \multicolumn{1}{c|}{\multirow{-2}{*}{53.91}} &
  \cellcolor[HTML]{EDEDE9}\textbf{6.95 ± 0.50} &
  \multicolumn{1}{c|}{\multirow{-2}{*}{53.33}} &
  \cellcolor[HTML]{EDEDE9}\textbf{4.86 ± 0.48} &
  \multicolumn{1}{c|}{\multirow{-2}{*}{57.11}} &
  \cellcolor[HTML]{EDEDE9}\textbf{6.66 ± 0.38} &
  \multicolumn{1}{c|}{\multirow{-2}{*}{46.40}} &
  \cellcolor[HTML]{EDEDE9}\textbf{7.07 ± 0.48} &
  \multirow{-2}{*}{53.82} \\ \midrule
\multicolumn{12}{c}{\cellcolor[HTML]{F5EBE0}Average} \\ \midrule
\multicolumn{1}{l|}{} &
  \multicolumn{1}{l|}{Binomial} &
  \multicolumn{1}{c}{12.76 ± 0.56} &
  \multicolumn{1}{c|}{} &
  \multicolumn{1}{c}{13.12 ± 0.49} &
  \multicolumn{1}{c|}{} &
  \multicolumn{1}{c}{14.98 ± 0.73} &
  \multicolumn{1}{c|}{} &
  \multicolumn{1}{c}{9.78 ± 0.29} &
  \multicolumn{1}{c|}{} &
  \multicolumn{1}{c}{10.91 ± 0.39} &
  \multicolumn{1}{c}{} \\
\multicolumn{1}{l|}{\multirow{-2}{*}{Average}} &
  \multicolumn{1}{l|}{\cellcolor[HTML]{EDEDE9}Ours} &
  \multicolumn{1}{c}{\cellcolor[HTML]{EDEDE9}\textbf{8.63 ± 0.60}} &
  \multicolumn{1}{c|}{\multirow{-2}{*}{49.73}} &
  \multicolumn{1}{c}{\cellcolor[HTML]{EDEDE9}\textbf{6.48 ± 0.51}} &
  \multicolumn{1}{c|}{\multirow{-2}{*}{51.81}} &
  \multicolumn{1}{c}{\cellcolor[HTML]{EDEDE9}\textbf{6.87 ± 0.48}} &
  \multicolumn{1}{c|}{\multirow{-2}{*}{52.86}} &
  \multicolumn{1}{c}{\cellcolor[HTML]{EDEDE9}\textbf{7.35 ± 0.42}} &
  \multicolumn{1}{c|}{\multirow{-2}{*}{43.14}} &
  \multicolumn{1}{c}{\cellcolor[HTML]{EDEDE9}\textbf{6.38 ± 0.44}} &
  \multicolumn{1}{c}{\multirow{-2}{*}{46.47}} \\ \bottomrule
\end{tabular}%
}
\vspace{-2.5mm}
\end{table*}

\begin{figure*}[!t]
    \centering
    \begin{minipage}[b]{0.48\textwidth}
        \centering
        \includegraphics[width=\linewidth]{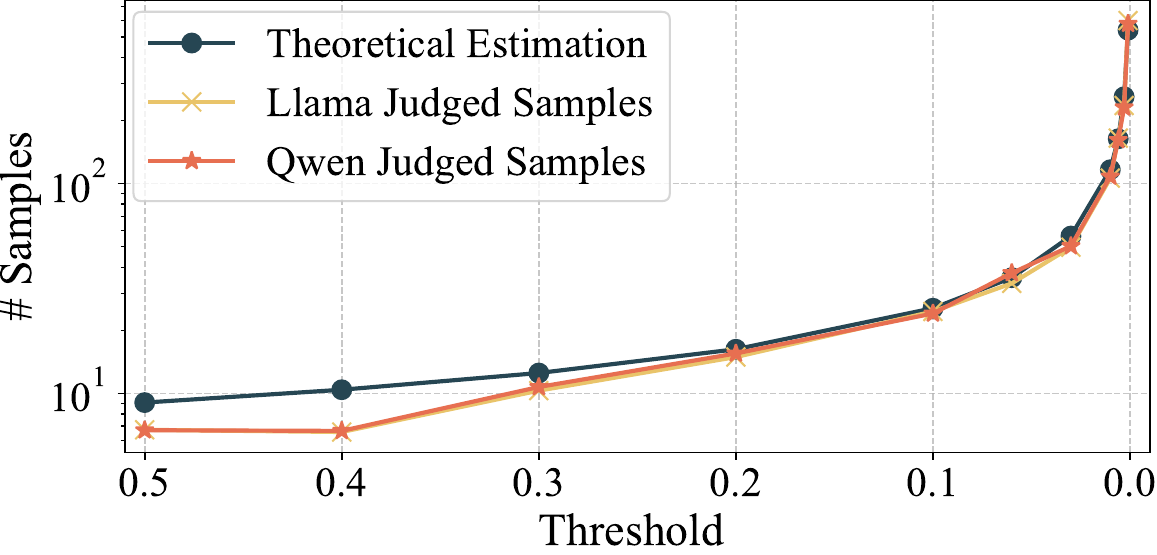}
        \vspace{-8mm}
        \caption{The actual number of samples under various thresholds $\xi$ versus the theoretical value from Equation (\ref{eq:sample_number}). \textbf{The actual sample numbers match with the theoretical bound.}}
        \label{fig:sampling_increase}
    \end{minipage}
    \hfill
    \begin{minipage}[b]{0.48\textwidth}
        \centering
        \includegraphics[width=\linewidth]{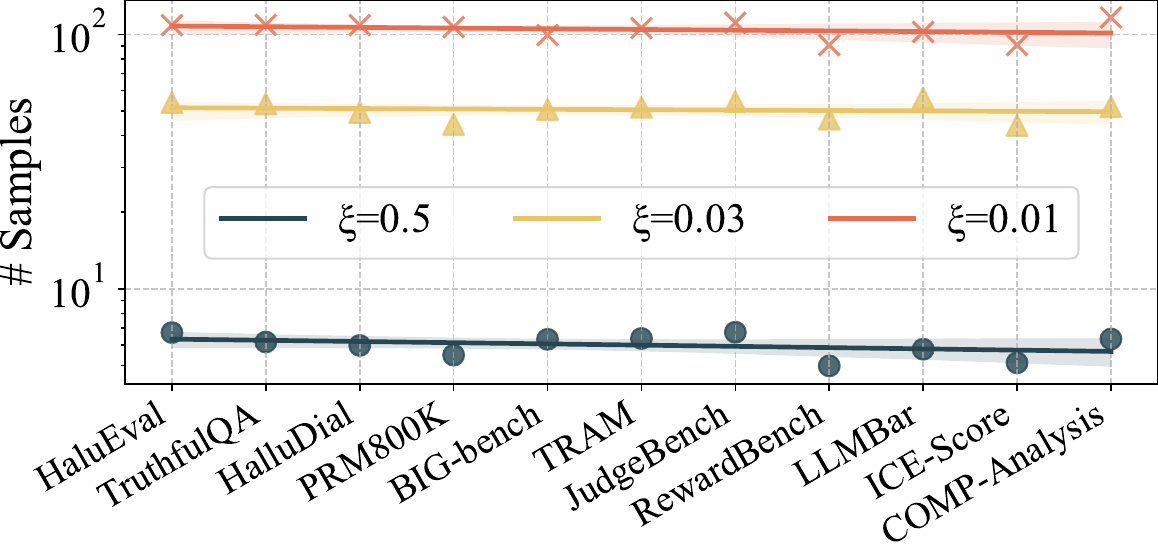}
        \vspace{-8mm}
        \caption{The actual number of samples from different datasets under three $\xi$ values. \textbf{Our sampling with adaptive stopping shows consistent results on all datasets.}}
        \label{fig:sample_datasets}
    \end{minipage}
    \vspace{-5mm}
\end{figure*}

\vspace{-3mm}
\subsection{Estimation Accuracy}
We begin by evaluating \bc~with adaptive stopping on datasets to verify its accuracy. We choose Binomial distribution and a single Beta-Binomial distribution as baselines and compare the error margin, which is the absolute difference between the estimation error rate and the actual value. The results of the error margin and sample numbers are reported in Table \ref{tab:est} and Table \ref{tab:single_beta}.

From the results, the following observations can be drawn: \ding{182} Compared to the Binomial distribution, \bc~achieves consistently lower error margin, with $32.4\% \sim 54.1\%$ improvements of average error margin of all models. This demonstrates an effective answer to \textbf{\hyperref[rq1]{RQ1}} by modeling judgment distribution as a mixture of Beta-Binomial distributions. \ding{183} The number of samples is close to the theoretical estimation. The average sample amount of models on all datasets exhibit a slight deviation of the estimated value $56$ by $3.14 \sim 12.86$ samples. This validates our design of using the distribution-free CP for adaptive stopping, which effectively solved \textbf{\hyperref[rq2]{RQ2}}.

\vspace{-3mm}
\subsection{Distribution Prior Transfer}
We then verify our text similarity-based distribution prior transfer when only limited samples are available. We constrain to $10$ samples from the target dataset and assume the full source datasets are accessible. Transfer is compared with estimating only on the target dataset samples (w/o Transfer). Error margins are shown in Table \ref{tab:transfer_result}.

From the results, we observe that by transferring from other datasets in the same category (e.g., from TruthfulQA and HalluDial to HaluEval), the average error margin across all datasets is reduced by $5.0\% \sim 25.0\%$ and is consistently lower compared to no transfer, suggesting the effectiveness of our design that using prior knowledge of the judgment distributions on open-source datasets can benefit estimation.
\vspace{-2.3em}

\begin{table*}[!t]
\caption{The comparison of error margins with and without distribution prior transfer. Estimations are performed using the mixture of Beta-Binomial distributions, with $10$ samples randomly drawn for evaluation. In experiments, each dataset is chosen as the target dataset, and the left datasets in the same domain are used as source datasets. \textbf{Bold} numbers denote lower mean margin. Scores are in percent (\%).}
\label{tab:transfer_result}
\centering
\resizebox{0.75\linewidth}{!}{%
\begin{tabular}{@{}llclclclclcl@{}}
\toprule
\multicolumn{1}{l|}{Dataset} &
  \multicolumn{1}{l|}{Method} &
  \multicolumn{2}{c}{Llama-3.3-70B} &
  \multicolumn{2}{c}{Qwen-2.5-72B} &
  \multicolumn{2}{c}{InternLM-2.5-20B} &
  \multicolumn{2}{c}{GPT-3.5} &
  \multicolumn{2}{c}{GPT-4} \\ \midrule
\multicolumn{12}{c}{\cellcolor[HTML]{F5EBE0}Hallucination Detection Datasets} \\ \midrule
\multicolumn{1}{l|}{} &
  \multicolumn{1}{l|}{w/o Transfer} &
  \multicolumn{2}{c}{12.43 ± 0.87} &
  \multicolumn{2}{c}{12.50 ± 0.92} &
  \multicolumn{2}{c}{10.09 ± 0.64} &
  \multicolumn{2}{c}{14.07 ± 0.75} &
  \multicolumn{2}{c}{12.85 ± 0.83} \\
\multicolumn{1}{l|}{\multirow{-2}{*}{HaluEval}} &
  \multicolumn{1}{l|}{\cellcolor[HTML]{EDEDE9}w/ Transfer} &
  \multicolumn{2}{c}{\cellcolor[HTML]{EDEDE9}\textbf{8.82 ± 0.42}} &
  \multicolumn{2}{c}{\cellcolor[HTML]{EDEDE9}\textbf{9.19 ± 0.75}} &
  \multicolumn{2}{c}{\cellcolor[HTML]{EDEDE9}\textbf{8.60 ± 0.64}} &
  \multicolumn{2}{c}{\cellcolor[HTML]{EDEDE9}\textbf{8.88 ± 0.71} }&
  \multicolumn{2}{c}{\cellcolor[HTML]{EDEDE9}\textbf{8.88 ± 0.86}} \\ \midrule
\multicolumn{1}{l|}{} &
  \multicolumn{1}{l|}{w/o Transfer} &
  \multicolumn{2}{c}{15.30 ± 0.81} &
  \multicolumn{2}{c}{13.88 ± 0.85} &
  \multicolumn{2}{c}{13.17 ± 1.11} &
  \multicolumn{2}{c}{12.54 ± 0.70	} &
  \multicolumn{2}{c}{13.21 ± 1.03	} \\
\multicolumn{1}{l|}{\multirow{-2}{*}{TruthfulQA}} &
  \multicolumn{1}{l|}{\cellcolor[HTML]{EDEDE9}w/ Transfer} &
  \multicolumn{2}{c}{\cellcolor[HTML]{EDEDE9}\textbf{3.37 ± 0.10}} &
  \multicolumn{2}{c}{\cellcolor[HTML]{EDEDE9}\textbf{8.55 ± 0.07}} &
  \multicolumn{2}{c}{\cellcolor[HTML]{EDEDE9}\textbf{10.18 ± 0.10}} &
  \multicolumn{2}{c}{\cellcolor[HTML]{EDEDE9}\textbf{10.18 ± 0.82	}} &
  \multicolumn{2}{c}{\cellcolor[HTML]{EDEDE9}\textbf{9.66 ± 0.70	}} \\ \midrule
\multicolumn{1}{l|}{} &
  \multicolumn{1}{l|}{w/o Transfer} &
  \multicolumn{2}{c}{17.53 ± 0.81} &
  \multicolumn{2}{c}{16.15 ± 0.60} &
  \multicolumn{2}{c}{11.35 ± 0.83} &
  \multicolumn{2}{c}{\textbf{16.62 ± 0.70}	} &
  \multicolumn{2}{c}{\textbf{14.64 ± 0.85}	} \\
\multicolumn{1}{l|}{\multirow{-2}{*}{HalluDial}} &
  \multicolumn{1}{l|}{\cellcolor[HTML]{EDEDE9}w/ Transfer} &
  \multicolumn{2}{c}{\cellcolor[HTML]{EDEDE9}\textbf{12.89 ± 0.77}} &
  \multicolumn{2}{c}{\cellcolor[HTML]{EDEDE9}\textbf{13.42 ± 0.53}} &
  \multicolumn{2}{c}{\cellcolor[HTML]{EDEDE9}\textbf{8.72 ± 0.54}} &
  \multicolumn{2}{c}{\cellcolor[HTML]{EDEDE9} 23.79 ± 0.84	} &
  \multicolumn{2}{c}{\cellcolor[HTML]{EDEDE9} 18.77 ± 0.92} \\ \midrule
\multicolumn{12}{c}{\cellcolor[HTML]{F5EBE0}Reasoning Datasets} \\ \midrule
\multicolumn{1}{l|}{} &
  \multicolumn{1}{l|}{w/o Transfer} &
  \multicolumn{2}{c}{\textbf{15.02 ± 0.78}} &
  \multicolumn{2}{c}{12.85 ± 0.88} &
  \multicolumn{2}{c}{\textbf{8.22 ± 0.58}} &
  \multicolumn{2}{c}{\textbf{9.27 ± 0.84	}} &
  \multicolumn{2}{c}{9.97 ± 0.53	} \\
\multicolumn{1}{l|}{\multirow{-2}{*}{PRM800K}} &
  \multicolumn{1}{l|}{\cellcolor[HTML]{EDEDE9}w/ Transfer} &
  \multicolumn{2}{c}{\cellcolor[HTML]{EDEDE9}15.11 ± 0.62} &
  \multicolumn{2}{c}{\cellcolor[HTML]{EDEDE9}\textbf{10.96 ± 0.99}} &
  \multicolumn{2}{c}{\cellcolor[HTML]{EDEDE9}8.46 ± 0.60} &
  \multicolumn{2}{c}{\cellcolor[HTML]{EDEDE9} 10.55 ± 0.84	} &
  \multicolumn{2}{c}{\cellcolor[HTML]{EDEDE9}\textbf{9.71 ± 1.00	}} \\ \midrule
\multicolumn{1}{l|}{} &
  \multicolumn{1}{l|}{w/o Transfer} &
  \multicolumn{2}{c}{15.22 ± 0.74} &
  \multicolumn{2}{c}{\textbf{13.81 ± 0.82}} &
  \multicolumn{2}{c}{\textbf{9.44 ± 0.53}} &
  \multicolumn{2}{c}{14.39 ± 0.74	} &
  \multicolumn{2}{c}{13.31 ± 1.15	} \\
\multicolumn{1}{l|}{\multirow{-2}{*}{BIG-bench}} &
  \multicolumn{1}{l|}{\cellcolor[HTML]{EDEDE9}w/ Transfer} &
  \multicolumn{2}{c}{\cellcolor[HTML]{EDEDE9}\textbf{12.69 ± 0.74}} &
  \multicolumn{2}{c}{\cellcolor[HTML]{EDEDE9}14.28 ± 0.79} &
  \multicolumn{2}{c}{\cellcolor[HTML]{EDEDE9}10.00 ± 0.62} &
  \multicolumn{2}{c}{\cellcolor[HTML]{EDEDE9}\textbf{9.98 ± 0.67	}} &
  \multicolumn{2}{c}{\cellcolor[HTML]{EDEDE9} \textbf{13.22 ± 0.69	}} \\ \midrule
\multicolumn{1}{l|}{} &
  \multicolumn{1}{l|}{w/o Transfer} &
  \multicolumn{2}{c}{14.77 ± 0.84} &
  \multicolumn{2}{c}{12.27 ± 0.69} &
  \multicolumn{2}{c}{11.67 ± 0.76} &
  \multicolumn{2}{c}{\textbf{13.52 ± 0.81	}} &
  \multicolumn{2}{c}{\textbf{12.69 ± 1.26	}} \\
\multicolumn{1}{l|}{\multirow{-2}{*}{TRAM}} &
  \multicolumn{1}{l|}{\cellcolor[HTML]{EDEDE9}w/ Transfer} &
  \multicolumn{2}{c}{\cellcolor[HTML]{EDEDE9}\textbf{12.52 ± 0.92}} &
  \multicolumn{2}{c}{\cellcolor[HTML]{EDEDE9}\textbf{11.03 ± 1.04}} &
  \multicolumn{2}{c}{\cellcolor[HTML]{EDEDE9}\textbf{10.85 ± 0.97}} &
  \multicolumn{2}{c}{\cellcolor[HTML]{EDEDE9}11.81 ± 1.00	} &
  \multicolumn{2}{c}{\cellcolor[HTML]{EDEDE9}11.25 ± 0.57	} \\ \midrule
\multicolumn{12}{c}{\cellcolor[HTML]{F5EBE0}Alignment Datasets} \\ \midrule
\multicolumn{1}{l|}{} &
  \multicolumn{1}{l|}{w/o Transfer} &
  \multicolumn{2}{c}{14.05 ± 0.88} &
  \multicolumn{2}{c}{12.41 ± 0.66} &
  \multicolumn{2}{c}{11.37 ± 0.79} &
  \multicolumn{2}{c}{\textbf{8.23 ± 0.75	}} &
  \multicolumn{2}{c}{\textbf{12.32 ± 0.69	}} \\
\multicolumn{1}{l|}{\multirow{-2}{*}{JudgeBench}} &
  \multicolumn{1}{l|}{\cellcolor[HTML]{EDEDE9}w/ Transfer} &
  \multicolumn{2}{c}{\cellcolor[HTML]{EDEDE9}\textbf{9.45 ± 0.59}} &
  \multicolumn{2}{c}{\cellcolor[HTML]{EDEDE9}\textbf{8.19 ± 0.66}} &
  \multicolumn{2}{c}{\cellcolor[HTML]{EDEDE9}\textbf{8.03 ± 0.54}} &
  \multicolumn{2}{c}{\cellcolor[HTML]{EDEDE9}14.36 ± 0.68	} &
  \multicolumn{2}{c}{\cellcolor[HTML]{EDEDE9}15.30 ± 1.19	} \\ \midrule
\multicolumn{1}{l|}{} &
  \multicolumn{1}{l|}{w/o Transfer} &
  \multicolumn{2}{c}{12.73 ± 0.68} &
  \multicolumn{2}{c}{\textbf{9.47 ± 1.07}} &
  \multicolumn{2}{c}{\textbf{10.34 ± 0.67}} &
  \multicolumn{2}{c}{\textbf{15.17 ± 0.92	}} &
  \multicolumn{2}{c}{13.30 ± 0.77	} \\
\multicolumn{1}{l|}{\multirow{-2}{*}{RewardBench}} &
  \multicolumn{1}{l|}{\cellcolor[HTML]{EDEDE9}w/ Transfer} &
  \multicolumn{2}{c}{\cellcolor[HTML]{EDEDE9}\textbf{12.72 ± 0.30}} &
  \multicolumn{2}{c}{\cellcolor[HTML]{EDEDE9}12.84 ± 0.48} &
  \multicolumn{2}{c}{\cellcolor[HTML]{EDEDE9}16.35 ± 0.36} &
  \multicolumn{2}{c}{\cellcolor[HTML]{EDEDE9}18.12 ± 0.34	} &
  \multicolumn{2}{c}{\cellcolor[HTML]{EDEDE9}\textbf{12.57 ± 0.38	}} \\ \midrule
\multicolumn{1}{l|}{} &
  \multicolumn{1}{l|}{w/o Transfer} &
  \multicolumn{2}{c}{16.97 ± 1.10} &
  \multicolumn{2}{c}{15.91 ± 0.70} &
  \multicolumn{2}{c}{10.03 ± 0.88} &
  \multicolumn{2}{c}{\textbf{17.00 ± 0.64	}} &
  \multicolumn{2}{c}{\textbf{12.90 ± 0.97	}} \\
\multicolumn{1}{l|}{\multirow{-2}{*}{LLMBar}} &
  \multicolumn{1}{l|}{\cellcolor[HTML]{EDEDE9}w/ Transfer} &
  \multicolumn{2}{c}{\cellcolor[HTML]{EDEDE9}\textbf{8.03 ± 0.39}} &
  \multicolumn{2}{c}{\cellcolor[HTML]{EDEDE9}\textbf{9.95 ± 0.30}} &
  \multicolumn{2}{c}{\cellcolor[HTML]{EDEDE9}\textbf{8.61 ± 0.41}} &
  \multicolumn{2}{c}{\cellcolor[HTML]{EDEDE9}21.94 ± 0.42} &
  \multicolumn{2}{c}{\cellcolor[HTML]{EDEDE9}17.70 ± 0.40	} \\ \midrule
\multicolumn{12}{c}{\cellcolor[HTML]{F5EBE0}Scoring Datasets} \\ \midrule
\multicolumn{1}{l|}{} &
  \multicolumn{1}{l|}{w/o Transfer} &
  \multicolumn{2}{c}{14.08 ± 0.53} &
  \multicolumn{2}{c}{\textbf{11.90 ± 1.05}} &
  \multicolumn{2}{c}{19.59 ± 0.78} &
  \multicolumn{2}{c}{12.11 ± 0.82	} &
  \multicolumn{2}{c}{13.98 ± 0.88	} \\
\multicolumn{1}{l|}{\multirow{-2}{*}{ICE-Score}} &
  \multicolumn{1}{l|}{\cellcolor[HTML]{EDEDE9}w/ Transfer} &
  \multicolumn{2}{c}{\cellcolor[HTML]{EDEDE9}\textbf{11.32 ± 0.66}} &
  \multicolumn{2}{c}{\cellcolor[HTML]{EDEDE9}11.99 ± 0.76} &
  \multicolumn{2}{c}{\cellcolor[HTML]{EDEDE9}\textbf{19.25 ± 1.05}} &
  \multicolumn{2}{c}{\cellcolor[HTML]{EDEDE9}\textbf{10.63 ± 0.66	}} &
  \multicolumn{2}{c}{\cellcolor[HTML]{EDEDE9}\textbf{12.30 ± 0.67	}} \\ \midrule
\multicolumn{1}{l|}{} &
  \multicolumn{1}{l|}{w/o Transfer} &
  \multicolumn{2}{c}{\textbf{14.85 ± 1.45}} &
  \multicolumn{2}{c}{\textbf{10.83 ± 0.60}} &
  \multicolumn{2}{c}{10.29 ± 0.60} &
  \multicolumn{2}{c}{10.22 ± 0.53	} &
  \multicolumn{2}{c}{16.18 ± 1.00	} \\
\multicolumn{1}{l|}{\multirow{-2}{*}{COMP-Analysis}} &
  \multicolumn{1}{l|}{\cellcolor[HTML]{EDEDE9}w/ Transfer} &
  \multicolumn{2}{c}{\cellcolor[HTML]{EDEDE9}15.29 ± 0.91} &
  \multicolumn{2}{c}{\cellcolor[HTML]{EDEDE9}12.28 ± 1.38} &
  \multicolumn{2}{c}{\cellcolor[HTML]{EDEDE9}\textbf{10.23 ± 0.72}} &
  \multicolumn{2}{c}{\cellcolor[HTML]{EDEDE9} \textbf{9.62 ± 0.53	}} &
  \multicolumn{2}{c}{\cellcolor[HTML]{EDEDE9}\textbf{14.97 ± 0.82	}} \\ \midrule
\multicolumn{12}{c}{\cellcolor[HTML]{F5EBE0}Average} \\ \midrule
\multicolumn{1}{l|}{} &
  \multicolumn{1}{l|}{w/o Transfer} &
  \multicolumn{2}{c}{14.81 ± 0.86} &
  \multicolumn{2}{c}{12.91 ± 0.80} &
  \multicolumn{2}{c}{11.41 ± 0.74} &
  \multicolumn{2}{c}{\textbf{13.01 ± 0.75}} &
  \multicolumn{2}{c}{13.21 ± 0.91	} \\
\multicolumn{1}{l|}{\multirow{-2}{*}{Average}} &
  \multicolumn{1}{l|}{\cellcolor[HTML]{EDEDE9}w/ Transfer} &
  \multicolumn{2}{c}{\cellcolor[HTML]{EDEDE9}\textbf{11.11 ± 0.58}} &
  \multicolumn{2}{c}{\cellcolor[HTML]{EDEDE9}\textbf{11.15 ± 0.70}} &
  \multicolumn{2}{c}{\cellcolor[HTML]{EDEDE9}\textbf{10.84 ± 0.60}} &
  \multicolumn{2}{c}{\cellcolor[HTML]{EDEDE9}13.62 ± 0.68}	 &
  \multicolumn{2}{c}{\cellcolor[HTML]{EDEDE9}	\textbf{13.12 ± 0.74	}} \\ \bottomrule
\end{tabular}%
}
\vspace{-4mm}
\end{table*}

\subsection{More Research Questions}

\paragraph{RQ3: Is sampling with adaptive stopping consistent to the theory?} We examine our adaptive stopping to see if Equation (\ref{eq:sample_number}) matches the real sampling amount. We set a series of $\xi$ while keeping $\tau=25$ and sample with adaptive stopping from judgment samples produced by Llama, Qwen, and GPT-4, and compare with the theoretical value of Equation (\ref{eq:sample_number}). The actual sample amounts under different thresholds in Figure \ref{fig:sampling_increase} match closely with the theoretical estimation, which proves the effectiveness of quantifying sampling deviation through CP and the Proposition \ref{pro:amount}.
\vspace{-4mm}

\paragraph{RQ4: Is adaptive stopping really distribution-free?} One benefit of adopting CP to quantify sampling deviation is distribution irrelevance. To testify to this, we consider sampling with various thresholds on all datasets to see if the sample amount remains consistent. The results in Figure \ref{fig:sample_datasets} show only a slight variance of sampling amounts across datasets, demonstrating superior stability. This verifies that our adaptive stopping is truly distribution-free, offering identical sampling guidance on diverse datasets.
\vspace{-4mm}

\paragraph{RQ5: Is CP-based Adaptive Stopping efficient?}
To validate the effectiveness of our CP-based adaptive stopping, we compare it against variance-based stopping. Specifically, we calculate the variance of sampling as
\begin{equation}
    \mathrm{Var}\left(sampling\right)=\frac{\alpha_r\beta_r}{(\alpha_r+\beta_r)^2(\alpha_r+\beta_r+1)},
\end{equation}
where $\alpha_r$ is the number of correct judgments in $r$ samples, and $\beta_r=r-\alpha_r$ is the number of wrong samples.

As shown in Table \ref{tab:cp_stopping}, CP consistently provides more effective guidance for adaptive stopping under the same deviation threshold $\xi$, which results in a reduced number of samples and achieves a reduction of up to 46.3\%.
\vspace{-3mm}

\begin{table}[!t]
\vspace{-3mm}
\caption{Comparison of variance-based adaptive stopping and our CP stopping. We compare the sample amount of both methods under the same threshold. \textbf{Bold} denotes less samples.}
\label{tab:cp_stopping}
\resizebox{\columnwidth}{!}{%
\begin{tabular}{@{}c|l|cccc@{}}
\toprule
                              &                           & HaluEval   & JudgeBench & PRM800K        & ICE-Score      \\ \cmidrule(l){3-6} 
\multirow{-2}{*}{Threshold $\xi$} & \multirow{-2}{*}{Methods} & \# Samples~($\downarrow$) & \# Samples~($\downarrow$) & \# Samples~($\downarrow$) & \# Samples~($\downarrow$) \\ \midrule
                              & Variance                  & 36.87      & 36.87      & \textbf{26.00} & \textbf{24.77} \\
\multirow{-2}{*}{$\xi$=0.06} &
  \cellcolor[HTML]{EDEDE9}Ours &
  \cellcolor[HTML]{EDEDE9}\textbf{35.37} &
  \cellcolor[HTML]{EDEDE9}\textbf{36.37} &
  \cellcolor[HTML]{EDEDE9}30.47 &
  \cellcolor[HTML]{EDEDE9}31.53 \\ \midrule
                              & Variance                  & 82.09      & 74.43      & 79.76          & 81.47          \\
\multirow{-2}{*}{$\xi$=0.03} &
  \cellcolor[HTML]{EDEDE9}Ours &
  \cellcolor[HTML]{EDEDE9}\textbf{54.72} &
  \cellcolor[HTML]{EDEDE9}\textbf{53.90} &
  \cellcolor[HTML]{EDEDE9}\textbf{43.32} &
  \cellcolor[HTML]{EDEDE9}\textbf{45.27} \\ \midrule
                              & Variance                  & 194.72     & 198.56     & 147.22         & 151.44         \\
\multirow{-2}{*}{$\xi$=0.01} &
  \cellcolor[HTML]{EDEDE9}Ours &
  \cellcolor[HTML]{EDEDE9}\textbf{109.06} &
  \cellcolor[HTML]{EDEDE9}\textbf{106.56} &
  \cellcolor[HTML]{EDEDE9}\textbf{101.28} &
  \cellcolor[HTML]{EDEDE9}\textbf{96.50} \\ \bottomrule
\end{tabular}%
}
\vspace{-5mm}
\end{table}

\section{Conclusion}
We present \bc, a framework for efficient estimation of LLM ensemble judge distribution. As part of our framework, we propose a mixture of Beta-Binomial distributions to model the judgment distribution after examining the inaccuracy of the Binomial assumption. We design conformal prediction-based adaptive stopping for sampling, which monitors the sampling deviation and effectively determines the sample amount for estimation. When only limited samples are available, we incorporate a text similarity-based distribution prior transfer mechanism to improve the estimation accuracy. As shown by experiments, the conformal prediction-based adaptive stopping effectively guided the sampling. Our mixture of Beta-Binomial distributions significantly outperforms the common Binomial assumption. With the transfer mechanism, \bc~can achieve high estimation precision with as few as $10$ samples.

\section*{Acknowledgement}
This project is supported by the Honda Research Institute USA. We extend our gratitude to Behzad Dariush for the insightful discussions. We also thank Zhaobo Zheng and David Isele for their valuable feedback and helpful suggestions.

%%%%%%%%%%%%%%%%%%%%%%%%%%%%%%%%%%%%%%%%%%%%%%%%%%%%%%%%%%%%%%%%%%%%%%%%%%%%%%%
%%%%%%%%%%%%%%%%%%%%%%%%%%%%%%%%%%%%%%%%%%%%%%%%%%%%%%%%%%%%%%%%%%%%%%%%%%%%%%%
% APPENDIX
%%%%%%%%%%%%%%%%%%%%%%%%%%%%%%%%%%%%%%%%%%%%%%%%%%%%%%%%%%%%%%%%%%%%%%%%%%%%%%%
%%%%%%%%%%%%%%%%%%%%%%%%%%%%%%%%%%%%%%%%%%%%%%%%%%%%%%%%%%%%%%%%%%%%%%%%%%%%%%%
% \section*{Impact Statement}
% This paper presents work whose goal is to advance the field of Machine Learning. There are many potential societal consequences of our work, none of which we feel must be specifically highlighted here.

\bibliography{example_paper}
\bibliographystyle{icml2025}

\newpage
\appendix
\onecolumn

\section{Proofs}

\subsection{Determination of Sample Amount.}
\label{app:stopping}
To derive a theoretical estimation of the sample amount for the adaptive stopping criteria above, we utilize the fundamental statistical properties of variance reduction with increasing sample size. Specifically, for i.i.d samples, the variance of the quantile decreases as:
\begin{equation}
    \mathrm{Var}(q_{1-\epsilon}^r) \propto \frac{1}{r\cdot f(q_{1-\epsilon})^2},
\end{equation}
where $f(q_{1-\epsilon})$ is the density function at the quantile.
The standard deviation of the estimator, which determines the variability of the quantile estimate, thus decays as:
\begin{equation}
    \mathrm{StdDev}(q_{1-\epsilon}^r) \propto \frac{1}{\sqrt{r}}.
\end{equation}
By the asymptotic theory of quantile estimation, for a large enough number of samples $r$, the empirical quantile $q^r_{1-\epsilon}$ converges to the quantile on the whole dataset $q_{1-\epsilon}$ with a known distribution based on Bahadur's representation:
\begin{equation}
    \sqrt{r} \left(q^r_{1-\epsilon} - q_{1-\epsilon}\right) \sim \mathcal{N} \left( 0, \frac{\epsilon(1-\epsilon)}{f(q_{1-\epsilon})^2} \right),
\end{equation}
This implies:
\begin{equation}
    q^r_{1-\epsilon} = q_{1-\epsilon}+O_p\left(\frac{1}{\sqrt{r}}\right),
\end{equation}
where $O_p(\cdot)$ denotes the order in probability. Thus, we can determine that the quantile itself decays as:
\begin{equation}
    \label{eq:cp_bigo}
    q_{1-\epsilon}^r-q_{1-\epsilon} = O_p\left(\frac{1}{\sqrt{r}}\right).
\end{equation}
This decay behavior shows that as $r$ increases, the estimated quantile approaches the theoretical quantile $q_{1-\epsilon}$, reflecting decreasing sampling deviation by using more samples. We will use this property to derive the relationship between the stopping criteria and the sample size $r$. From the stopping criteria in Equation (\ref{eq:stop_criteria}), 
\begin{equation}
    \left|q_{1-\epsilon}^r-q_{1-\epsilon}^{r-1}\right| \leq \xi.
\end{equation}
According to the calculations in Equation (\ref{eq:cp_bigo}), we can rewrite the bound for $q_{1-\epsilon}^{r-1}$ as
\begin{equation}
    \label{eq:cp_bigO_r}
    q_{1-\epsilon}^{r-1}-q_{1-\epsilon}=O_p\left(\frac{1}{\sqrt{r-1}}\right).
\end{equation}
Thus we have
\begin{equation}
    \label{eq:triangle}
    \left|q_{1-\epsilon}^r-q_{1-\epsilon}^{r-1}\right| = O_p\left(\frac{1}{\sqrt{r}}-\frac{1}{\sqrt{r-1}}\right).
\end{equation}
This suggests to meet Equation (\ref{eq:stop_criteria}), it requires
\begin{equation}
    \tau\left(\frac{1}{\sqrt{r-1}}-\frac{1}{\sqrt{r}}\right)<\xi,
\end{equation}
which proves Equation (\ref{eq:sample_number}).

\subsection{Error Rate with Adaptive Sampling}
\label{app: error}
In this section we develop a theoretical estimation of the error bound for adaptive sampling. We first consider the base case and as shown in Equation (\ref{eq:BB_error}), we know that the mixture distribution error rate is: 
\begin{equation}
        \Tilde{P}_\mathrm{BB} = w\sum_{s=0}^{\lceil k/2 \rceil-1} \binom{k}{s}\frac{\mathrm{B}(s+\alpha_1,k-s+\beta_1)}{\mathrm{B}(\alpha_1,\beta_1)} + (1-w)\sum_{s=0}^{\lceil k/2 \rceil-1} \binom{k}{s}\frac{\mathrm{B}(s+\alpha_2,k-s+\beta_2)}{\mathrm{B}(\alpha_2,\beta_2)}
\end{equation}
The adaptive stopping criterion is given by Equation (\ref{eq:stop_criteria}):
% \begin{equation}
%     \left|\frac{q_{1-\epsilon}^r - q_{1-\epsilon}^{r-1}}{q_{1-\epsilon}^{r-1}}\right| \leq \xi
% \end{equation}
\begin{equation}
  \left|q_{1-\epsilon}^r-q_{1-\epsilon}^{r-1}\right| \leq \xi.
\end{equation}
The sample size requirement is given by Equation (\ref{eq:sample_number}): 
\begin{equation}
    \tau\left(\frac{1}{\sqrt{r-1}}-\frac{1}{\sqrt{r}}\right) \leq \xi.
\end{equation}
Based on the two equations and large number theory, we know that the difference between the quantile on samples $q^r_{1-\epsilon}$ and the quantile on the whole dataset $q_{1-\epsilon}$ decays proportionally to $\frac{\tau}{\sqrt{r}}$. In addition, the non-conformity score $s_i$ is defined in Equation (\ref{eq:nonconformity_score}):
\begin{equation}
    s_i=\text{score}(S_i) = |S_i - \mathbb{E}[S]|,
\end{equation}
 where $S_i$ is the number of correct judgments in the $i$-th sample. As the $(1-\epsilon)$-quantile of the sorted scores $s_1<\ldots<s_r$ at stopping time with $r$ samples is:
\begin{equation}
    q_{1-\epsilon}^r = s_{\lceil(1-\epsilon)\cdot(r+1)\rceil}.
\end{equation}
% From Equation 10, we know that the quantile $q_{1-\epsilon}^r $ decays as:
% \begin{equation}
% q_{1-\epsilon}^r \propto \frac{\tau}{\sqrt{r}}
% \end{equation}
% % From Equation 10, our nonconformity scores are:
% %     \begin{equation}
% %     \text{score}(S_i) = |S_i - \mathbb{E}[S]|
% %     \end{equation}
% %     where $S_i$ is the number of correct judgments in the $i$-th sample.
% %  The $(1-\epsilon)$-quantile of these scores at stopping time $r$ is:
% %     \begin{equation}
% %     q_{1-\epsilon}^r = \text{score}(S_{\lceil(1-\epsilon)\cdot(r+1)\rceil})
% %     \end{equation}
 When the stopping criterion is met, this implies the confidence region for $\mathbb{E}[S]$ has stabilized and the following holds:
    \begin{equation}
    \mathbb{P}(|S_i - \mathbb{E}[S]| \leq q_{1-\epsilon}^r) = 1-\epsilon.
    \end{equation}

For the Beta-Binomial mixture model, $\mathbb{E}[S]$ relates to the error rate via:
\begin{equation}
    \Tilde{P}_\mathrm{BB}= \mathbb{P}(S < \lceil k/2 \rceil).
\end{equation}
We will use the quantile stability argument as follows. For a sequence of independent samples $\{S_1, ..., S_r\}$, let $s_i$ be the non-conformity score defined as:
\begin{equation}
    s_i=\text{score}(S_i) = |S_i - \mathbb{E}[S]|,
\end{equation}
where $S_i$ is the number of correct judgments in the $i$-th sample. By the theory of quantile estimation, for a large enough number of samples $r$, the empirical quantile $q_{1-\epsilon}^r$ converges to the population quantile $q_{1-\epsilon}$ with a known distribution:
\begin{equation}
    \sqrt{r}(q_{1-\epsilon}^r - q_{1-\epsilon}) \sim \mathcal{N}\left(0, \frac{\epsilon(1-\epsilon)}{f(q_{1-\epsilon})^2}\right),
\end{equation}
where $f(\cdot)$ is the density function. This implies:
\begin{equation}
    q_{1-\epsilon}^r = q_{1-\epsilon} + O_p\left(\frac{1}{\sqrt{r}}\right),
\end{equation}
where $O_p(\cdot)$ denotes the order in probability. As the $(1-\epsilon)$-quantile of the sorted scores $s_1<\ldots<s_r$ at stopping time with $r$ samples is:
\begin{equation}
    q_{1-\epsilon}^r = s_{\lceil(1-\epsilon)\cdot(r+1)\rceil}.
\end{equation}
When the stopping criterion is met, this implies the confidence region for $\mathbb{E}[S]$ has stabilized and the following holds:
\begin{equation}
    \mathbb{P}(|S_i - \mathbb{E}[S]| \leq q_{1-\epsilon}^r) = 1-\epsilon.
\end{equation}
For the Beta-Binomial mixture model, $\mathbb{E}[S]$ relates to the error rate via:
\begin{equation}
    \Tilde{P}_\mathrm{BB}= \mathbb{P}(S < \lceil k/2 \rceil).
\end{equation}
By the quantile stability argument above, we have the bound:
\begin{equation}
    (1 - \min(\xi, \frac{\tau}{\sqrt{r}}))\mathbb{E}[S]_\mathrm{BB} <  \mathbb{E}[S]_{adapt} <  (1 + \min(\xi, \frac{\tau}{\sqrt{r}}))\mathbb{E}[S]_\mathrm{BB}
\end{equation}
The error probability of $\Tilde{P}_\mathrm{BB}$ is defined using the Beta-Binomial cumulative distribution function:
\begin{equation}
    \Tilde{P}_\mathrm{BB} = \mathbb{P}(S < \lceil k/2 \rceil) = F_\mathrm{BB}(\lceil k/2 \rceil - 1),
\end{equation}
where $F_\mathrm{BB}$ is the Beta-Binomial cumulative distribution function. Since $F_\mathrm{BB}$  is monotonically increasing, the error probability $\Tilde{P}_{adapt}$ follows the same proportional bound. 

% Given the confidence bounds on $\mathbb{E}[S]$ from the quantile stability and the monotonicity of the Beta-Binomial cumulative distribution function:
\begin{equation}
    ( 1 - \min(\xi, \frac{\tau}{\sqrt{r}}))\Tilde{P}_\mathrm{BB}
    < \Tilde{P}_{adpt} <  ( 1 + \min(\xi, \frac{\tau}{\sqrt{r}}))\Tilde{P}_\mathrm{BB}.
\end{equation}
Therefore, we have:
    \begin{equation}
   \Tilde{P}_{adapt} = ( 1 \pm \min(\xi, \frac{\tau}{\sqrt{r}}))\Tilde{P}_\mathrm{BB}.
    \end{equation}
%     \end{equation}

\section{Implementation Details}
In this section, we elaborate on the implementation details of \bc. 

We evaluate LLM ensembles of $k\in \{1, 3, 5, 7, 9, 11 \}$ models, including GPT-3.5 \citep{brown2020language}, GPT-4 \citep{openai2024gpt4technicalreport}), Llama-3.3-70B \citep{dubey2024llama}, Qwen-2.5-72B \citep{yang2024qwen2} and InternLM-2.5-20B \citep{cai2024internlm2}). We choose domains of \textit{hallucination detection} (HaluEval, \citealp{li2023halueval}; TruthfulQA, \citealp{lin2021truthfulqa}; HalluDial, \citealp{luo2024halludial}), \textit{reasoning} (PRM800K, \citealp{lightman2023let}; BIG-bench, \citealp{srivastava2022beyond}; TRAM, \citealp{wang2023tram}), \textit{scoring} (ICE-Score, \citealp{zhuo2023ice}; Comp-Analysis, \citealp{zhang2024comprehensive}) and \textit{alignment} (JudgeBench, \citealp{tan2024judgebench}; RewardBench, \citealp{lambert2024rewardbench}; LLMBar, \citealp{zeng2023evaluating}).

Throughout all the experiments, the sampling temperature of all LLMs is set to $1$. We do not set random seeds for all experiments. All experiments are run for $30$ times to calculate the mean and standard deviation of the error margin.

\section{Additional Experiments}
In this section, we conduct additional experiments. Specifically, in Table \ref{tab:single_beta}, we compare \bc~with a single Beta-Binomial distribution. Results demonstrate that we achieve superior performance due to the incorporation of a mixture of Beta-Binomial distributions. In Table \ref{tab:ablation}, we conduct ablation studies on our distribution transfer design. Compared to ablated variants, our full design achieves the smallest error margin, indicating that our transfer design is effective.

\begin{table*}[t]
\caption{The comparison of error margins between our \textbf{mixture of Beta-Binomial} distributions and a \textbf{single Beta-Binomial} distribution. The error margin is calculated as the absolute difference between the actual error rate and the estimation. Estimations using both distributions are done on samples obtained by $\texttt{BetaConform}$ through iterative sampling with adaptive stopping. For each run, the error margin is computed from $k=1$ to $11$, and the average margin of ensemble sizes is used as the result for that run. We conduct $30$ runs and report the average and standard deviation. The average number of samples across runs is also reported.}
\label{tab:single_beta}
\resizebox{\textwidth}{!}{%
\begin{tabular}{@{}llcccccc@{}}
\toprule
\multicolumn{1}{l|}{} &
  \multicolumn{1}{l|}{} &
  \multicolumn{2}{c|}{Llama-3.3-70B} &
  \multicolumn{2}{c|}{Qwen-2.5-72B} &
  \multicolumn{2}{c}{InternLM-20B} \\ \cmidrule(l){3-8} 
\multicolumn{1}{l|}{\multirow{-2}{*}{Dataset}} &
  \multicolumn{1}{l|}{\multirow{-2}{*}{Method}} &
  Error Margin $(\downarrow)$ &
  \multicolumn{1}{c|}{\# Samples $(\downarrow)$} &
  Error Margin $(\downarrow)$ &
  \multicolumn{1}{c|}{\# Samples $(\downarrow)$} &
  Error Margin $(\downarrow)$ &
  \# Samples $(\downarrow)$ \\ \midrule
\multicolumn{8}{c}{\cellcolor[HTML]{F5EBE0}Hallucination Detection} \\ \midrule
\multicolumn{1}{l|}{} &
  \multicolumn{1}{l|}{Beta-Binomial} &
  14.46 ± 0.16 &
  \multicolumn{1}{c|}{} &
  5.14 ± 0.21 &
  \multicolumn{1}{c|}{} &
  15.92 ± 0.11 &
   \\
\multicolumn{1}{l|}{\multirow{-2}{*}{HaluEval}} &
  \multicolumn{1}{l|}{\cellcolor[HTML]{EDEDE9}Ours} &
  \cellcolor[HTML]{EDEDE9}\textbf{6.68 ± 0.53} &
  \multicolumn{1}{c|}{\multirow{-2}{*}{49.47}} &
  \cellcolor[HTML]{EDEDE9}\textbf{4.72 ± 0.38} &
  \multicolumn{1}{c|}{\multirow{-2}{*}{61.02}} &
  \cellcolor[HTML]{EDEDE9}\textbf{5.48 ± 0.41} &
  \multirow{-2}{*}{50.67} \\ \midrule
\multicolumn{1}{l|}{} &
  \multicolumn{1}{l|}{Beta-Binomial} &
  8.83 ± 1.02 &
  \multicolumn{1}{c|}{} &
  7.84 ± 0.26 &
  \multicolumn{1}{c|}{} &
  6.79 ± 0.25 &
   \\
\multicolumn{1}{l|}{\multirow{-2}{*}{TruthfulQA}} &
  \multicolumn{1}{l|}{\cellcolor[HTML]{EDEDE9}Ours} &
  \cellcolor[HTML]{EDEDE9}\textbf{7.53 ± 0.55} &
  \multicolumn{1}{c|}{\multirow{-2}{*}{54.13}} &
  \cellcolor[HTML]{EDEDE9}\textbf{7.18 ± 0.44} &
  \multicolumn{1}{c|}{\multirow{-2}{*}{53.56}} &
  \cellcolor[HTML]{EDEDE9}\textbf{6.24 ± 0.59} &
  \multirow{-2}{*}{55.56} \\ \midrule
\multicolumn{1}{l|}{} &
  \multicolumn{1}{l|}{Beta-Binomial} &
  11.33 ± 0.64 &
  \multicolumn{1}{c|}{} &
  16.75 ± 0.90 &
  \multicolumn{1}{c|}{} &
  7.95 ± 0.34 &
   \\
\multicolumn{1}{l|}{\multirow{-2}{*}{HalluDial}} &
  \multicolumn{1}{l|}{\cellcolor[HTML]{EDEDE9}Ours} &
  \cellcolor[HTML]{EDEDE9}\textbf{7.94 ± 0.68} &
  \multicolumn{1}{c|}{\multirow{-2}{*}{46.58}} &
  \cellcolor[HTML]{EDEDE9}\textbf{6.96 ± 0.47} &
  \multicolumn{1}{c|}{\multirow{-2}{*}{55.78}} &
  \cellcolor[HTML]{EDEDE9}\textbf{6.43 ± 0.50} &
  \multirow{-2}{*}{51.87} \\ \midrule
\multicolumn{8}{c}{\cellcolor[HTML]{F5EBE0}Reasoning} \\ \midrule
\multicolumn{1}{l|}{} &
  \multicolumn{1}{l|}{Beta-Binomial} &
  16.45 ± 1.35 &
  \multicolumn{1}{c|}{} &
  10.30 ± 0.60 &
  \multicolumn{1}{c|}{} &
  9.81 ± 0.61 &
   \\
\multicolumn{1}{l|}{\multirow{-2}{*}{PRM800K}} &
  \multicolumn{1}{l|}{\cellcolor[HTML]{EDEDE9}Ours} &
  \cellcolor[HTML]{EDEDE9}\textbf{9.37 ± 0.64} &
  \multicolumn{1}{c|}{\multirow{-2}{*}{43.33}} &
  \cellcolor[HTML]{EDEDE9}\textbf{7.82 ± 0.69} &
  \multicolumn{1}{c|}{\multirow{-2}{*}{42.89}} &
  \cellcolor[HTML]{EDEDE9}\textbf{4.52 ± 0.50} &
  \multirow{-2}{*}{46.13} \\ \midrule
\multicolumn{1}{l|}{} &
  \multicolumn{1}{l|}{Beta-Binomial} &
  13.15 ± 0.68 &
  \multicolumn{1}{c|}{} &
  12.32 ± 0.60 &
  \multicolumn{1}{c|}{} &
  9.51 ± 0.56 &
   \\
\multicolumn{1}{l|}{\multirow{-2}{*}{BIG-bench}} &
  \multicolumn{1}{l|}{\cellcolor[HTML]{EDEDE9}Ours} &
  \cellcolor[HTML]{EDEDE9}\textbf{11.15 ± 0.60} &
  \multicolumn{1}{c|}{\multirow{-2}{*}{51.51}} &
  \cellcolor[HTML]{EDEDE9}\textbf{6.97 ± 0.58} &
  \multicolumn{1}{c|}{\multirow{-2}{*}{47.82}} &
  \cellcolor[HTML]{EDEDE9}\textbf{5.54 ± 0.51} &
  \multirow{-2}{*}{48.40} \\ \midrule
\multicolumn{1}{l|}{} &
  \multicolumn{1}{l|}{Beta-Binomial} &
  11.75 ± 0.74 &
  \multicolumn{1}{c|}{} &
  \textbf{5.72 ± 0.39} &
  \multicolumn{1}{c|}{} &
  \textbf{6.01 ± 0.44} &
   \\
\multicolumn{1}{l|}{\multirow{-2}{*}{TRAM}} &
  \multicolumn{1}{l|}{\cellcolor[HTML]{EDEDE9}Ours} &
  \cellcolor[HTML]{EDEDE9}\textbf{8.39 ± 0.63} &
  \multicolumn{1}{c|}{\multirow{-2}{*}{55.87}} &
  \cellcolor[HTML]{EDEDE9}6.20 ± 0.34 &
  \multicolumn{1}{c|}{\multirow{-2}{*}{57.16}} &
  \cellcolor[HTML]{EDEDE9}6.10 ± 0.58 &
  \multirow{-2}{*}{57.78} \\ \midrule
\multicolumn{8}{c}{\cellcolor[HTML]{F5EBE0}Alignment} \\ \midrule
\multicolumn{1}{l|}{} &
  \multicolumn{1}{l|}{Beta-Binomial} &
  7.60 ± 0.37 &
  \multicolumn{1}{c|}{} &
  7.64 ± 0.54 &
  \multicolumn{1}{c|}{} &
  \textbf{5.11 ± 0.24} &
   \\
\multicolumn{1}{l|}{\multirow{-2}{*}{JudgeBench}} &
  \multicolumn{1}{l|}{\cellcolor[HTML]{EDEDE9}Ours} &
  \cellcolor[HTML]{EDEDE9}\textbf{6.98 ± 0.56} &
  \multicolumn{1}{c|}{\multirow{-2}{*}{60.58}} &
  \cellcolor[HTML]{EDEDE9}\textbf{5.39 ± 0.39} &
  \multicolumn{1}{c|}{\multirow{-2}{*}{58.40}} &
  \cellcolor[HTML]{EDEDE9}5.26 ± 0.39 &
  \multirow{-2}{*}{57.16} \\ \midrule
\multicolumn{1}{l|}{} &
  \multicolumn{1}{l|}{Beta-Binomial} &
  16.29 ± 1.39 &
  \multicolumn{1}{c|}{} &
  11.40 ± 1.20 &
  \multicolumn{1}{c|}{} &
  \textbf{6.15 ± 0.27} &
   \\
\multicolumn{1}{l|}{\multirow{-2}{*}{RewardBench}} &
  \multicolumn{1}{l|}{\cellcolor[HTML]{EDEDE9}Ours} &
  \cellcolor[HTML]{EDEDE9}\textbf{11.30 ± 0.62} &
  \multicolumn{1}{c|}{\multirow{-2}{*}{40.22}} &
  \cellcolor[HTML]{EDEDE9}\textbf{4.68 ± 0.56} &
  \multicolumn{1}{c|}{\multirow{-2}{*}{45.20}} &
  \cellcolor[HTML]{EDEDE9}6.58 ± 0.40 &
  \multirow{-2}{*}{52.04} \\ \midrule
\multicolumn{1}{l|}{} &
  \multicolumn{1}{l|}{Beta-Binomial} &
  14.21 ± 0.67 &
  \multicolumn{1}{c|}{} &
  7.97 ± 0.58 &
  \multicolumn{1}{c|}{} &
  \textbf{5.46 ± 0.30} &
   \\
\multicolumn{1}{l|}{\multirow{-2}{*}{LLMBar}} &
  \multicolumn{1}{l|}{\cellcolor[HTML]{EDEDE9}Ours} &
  \cellcolor[HTML]{EDEDE9}\textbf{10.18 ± 0.71} &
  \multicolumn{1}{c|}{\multirow{-2}{*}{50.18}} &
  \cellcolor[HTML]{EDEDE9}\textbf{7.52 ± 0.63} &
  \multicolumn{1}{c|}{\multirow{-2}{*}{51.07}} &
  \cellcolor[HTML]{EDEDE9}6.38 ± 0.53 &
  \multirow{-2}{*}{51.29} \\ \midrule
\multicolumn{8}{c}{\cellcolor[HTML]{F5EBE0}Scoring} \\ \midrule
\multicolumn{1}{l|}{} &
  \multicolumn{1}{l|}{Beta-Binomial} &
  16.71 ± 1.11 &
  \multicolumn{1}{c|}{} &
  9.24 ± 0.59 &
  \multicolumn{1}{c|}{} &
  \textbf{10.97 ± 0.27} &
   \\
\multicolumn{1}{l|}{\multirow{-2}{*}{ICE-Score}} &
  \multicolumn{1}{l|}{\cellcolor[HTML]{EDEDE9}Ours} &
  \cellcolor[HTML]{EDEDE9}\textbf{8.97 ± 0.45} &
  \multicolumn{1}{c|}{\multirow{-2}{*}{41.29}} &
  \cellcolor[HTML]{EDEDE9}\textbf{6.91 ± 0.59} &
  \multicolumn{1}{c|}{\multirow{-2}{*}{43.73}} &
  \cellcolor[HTML]{EDEDE9}18.19 ± 0.37 &
  \multirow{-2}{*}{53.42} \\ \midrule
\multicolumn{1}{l|}{} &
  \multicolumn{1}{l|}{Beta-Binomial} &
  8.56 ± 0.66 &
  \multicolumn{1}{c|}{} &
  \textbf{6.93 ± 0.34} &
  \multicolumn{1}{c|}{} &
  \textbf{4.61 ± 0.27} &
   \\
\multicolumn{1}{l|}{\multirow{-2}{*}{COMP-Analysis}} &
  \multicolumn{1}{l|}{\cellcolor[HTML]{EDEDE9}Ours} &
  \cellcolor[HTML]{EDEDE9}\textbf{6.50 ± 0.63} &
  \multicolumn{1}{c|}{\multirow{-2}{*}{53.91}} &
  \cellcolor[HTML]{EDEDE9}6.95 ± 0.50 &
  \multicolumn{1}{c|}{\multirow{-2}{*}{53.33}} &
  \cellcolor[HTML]{EDEDE9}4.86 ± 0.48 &
  \multirow{-2}{*}{57.11} \\ \midrule
\multicolumn{8}{c}{\cellcolor[HTML]{F5EBE0}Average} \\ \midrule
\multicolumn{1}{l|}{} &
  \multicolumn{1}{l|}{Beta-Binomial} &
  12.67 ± 0.80 &
  \multicolumn{1}{c|}{} &
  9.20 ± 0.56 &
  \multicolumn{1}{c|}{} &
  8.03 ± 0.33 &
   \\
\multicolumn{1}{l|}{\multirow{-2}{*}{Average}} &
  \multicolumn{1}{l|}{\cellcolor[HTML]{EDEDE9}Ours} &
  \cellcolor[HTML]{EDEDE9}\textbf{8.63 ± 0.60} &
  \multicolumn{1}{c|}{\multirow{-2}{*}{49.73}} &
  \cellcolor[HTML]{EDEDE9}\textbf{6.48 ± 0.51} &
  \multicolumn{1}{c|}{\multirow{-2}{*}{51.81}} &
  \cellcolor[HTML]{EDEDE9}\textbf{6.87 ± 0.48} &
  \multirow{-2}{*}{52.86} \\ \bottomrule
\end{tabular}%
}
\end{table*}

\clearpage

\begin{table*}[t]
\caption{The ablation study of \texttt{BetaConform} distribution prior transfer. \ding{182} $\log(r_i) \rightarrow r_i$ means the first term $\log(r_i)$ in Eq. 14 is replaced with $r_i$ to still asign a larger dataset higher weight while not considering source datasets could be magnitudes larger. \ding{183} $\mathrm{CosSim}(\Bar{E}_0, \Bar{E}_i) \rightarrow \frac{1}{|\bar{E}_0 - \bar{E}_i|_2}$ refers to replacing the cosine similarity to measure the source datasets and the target dataset with the reciprocal of the Euclidean distance between the embeddings of the two datasets. This still assigns more similar datasets higher weights. \ding{184} No $\sigma(\cdot)$ means the transfer weight is computed as $\lambda_i=\log(r_i)\cdot\mathrm{CosSim}(\Bar{E}_0, \Bar{E}_i)$, without using the sigmoid function $\sigma(\cdot)$ to reduce the weight of low similarity datasets}
\label{tab:ablation}
\centering
\resizebox{0.75\textwidth}{!}{%
\begin{tabular}{@{}l|l|c|c|c@{}}
\toprule
 &
   &
  Llama-3.3-70B &
  Qwen-2.5-72B &
  InternLM-20B \\ \cmidrule(l){3-5} 
\multirow{-2}{*}{Dataset} &
  \multirow{-2}{*}{Ablation} &
  Error Margin &
  Error Margin &
  Error Margin \\ \midrule
 &
  $\log(r_i) \rightarrow r_i$ &
  10.94 ± 0.57 &
  9.53 ± 0.70 &
  11.16 ± 0.75 \\ \cmidrule(l){2-5} 
 &
  $\mathrm{CosSim}(\Bar{E}_0, \Bar{E}_i) \rightarrow \frac{1}{|\bar{E}_0 - \bar{E}_i|_2}$ &
  11.90 ± 0.85 &
  13.17 ± 0.68 &
  10.29 ± 0.80 \\ \cmidrule(l){2-5} 
 &
  No $\sigma(\cdot)$ &
  10.04 ± 0.23 &
  23.03 ± 0.12 &
  \textbf{8.45 ± 0.10} \\ \cmidrule(l){2-5} 
\multirow{-4}{*}{HaluEval} &
  \cellcolor[HTML]{EDEDE9}Ours &
  \cellcolor[HTML]{EDEDE9}\textbf{8.82 ± 0.42} &
  \cellcolor[HTML]{EDEDE9}\textbf{9.19 ± 0.75} &
  \cellcolor[HTML]{EDEDE9}8.60 ± 0.64 \\ \midrule
 &
  $\log(r_i) \rightarrow r_i$ &
  13.47 ± 0.66 &
  11.17 ± 1.15 &
  10.65 ± 0.89 \\ \cmidrule(l){2-5} 
 &
  $\mathrm{CosSim}(\Bar{E}_0, \Bar{E}_i) \rightarrow \frac{1}{|\bar{E}_0 - \bar{E}_i|_2}$ &
  15.13 ± 0.71 &
  13.14 ± 0.96 &
  11.03 ± 0.80 \\ \cmidrule(l){2-5} 
 &
  No $\sigma(\cdot)$ &
  6.87 ± 0.01 &
  16.52 ± 0.03 &
  12.47 ± 0.06 \\ \cmidrule(l){2-5} 
\multirow{-4}{*}{TruthfulQA} &
  \cellcolor[HTML]{EDEDE9}Ours &
  \cellcolor[HTML]{EDEDE9}\textbf{3.37 ± 0.10} &
  \cellcolor[HTML]{EDEDE9}\textbf{8.55 ± 0.07} &
  \cellcolor[HTML]{EDEDE9}\textbf{10.18 ± 0.10} \\ \midrule
 &
  $\log(r_i) \rightarrow r_i$ &
  13.55 ± 0.58 &
  15.43 ± 0.86 &
  10.42 ± 1.00 \\ \cmidrule(l){2-5} 
 &
  $\mathrm{CosSim}(\Bar{E}_0, \Bar{E}_i) \rightarrow \frac{1}{|\bar{E}_0 - \bar{E}_i|_2}$ &
  15.54 ± 0.59 &
  15.89 ± 0.65 &
  10.47 ± 0.67j \\ \cmidrule(l){2-5} 
 &
  No $\sigma(\cdot)$ &
  \textbf{12.39 ± 0.00} &
  16.61 ± 0.09 &
  13.00 ± 0.07 \\ \cmidrule(l){2-5} 
\multirow{-4}{*}{HalluDial} &
  \cellcolor[HTML]{EDEDE9}Ours &
  \cellcolor[HTML]{EDEDE9}12.89 ± 0.77 &
  \cellcolor[HTML]{EDEDE9}\textbf{13.42 ± 0.53} &
  \cellcolor[HTML]{EDEDE9}\textbf{8.72 ± 0.54} \\ \midrule
 &
  $\log(r_i) \rightarrow r_i$ &
  25.97 ± 0.03 &
  21.23 ± 0.04 &
  15.46 ± 0.06 \\ \cmidrule(l){2-5} 
 &
  $\mathrm{CosSim}(\Bar{E}_0, \Bar{E}_i) \rightarrow \frac{1}{|\bar{E}_0 - \bar{E}_i|_2}$ &
  14.43 ± 1.12 &
  11.26 ± 0.99 &
  11.47 ± 0.74 \\ \cmidrule(l){2-5} 
 &
  No $\sigma(\cdot)$ &
  24.57 ± 0.44 &
  19.26 ± 0.13 &
  10.42 ± 0.08 \\ \cmidrule(l){2-5} 
\multirow{-4}{*}{JudgeBench} &
  \cellcolor[HTML]{EDEDE9}Ours &
  \cellcolor[HTML]{EDEDE9}\textbf{9.45 ± 0.59} &
  \cellcolor[HTML]{EDEDE9}\textbf{8.19 ± 0.66} &
  \cellcolor[HTML]{EDEDE9}\textbf{8.03 ± 0.54} \\ \midrule
 &
  $\log(r_i) \rightarrow r_i$ &
  15.00 ± 0.01 &
  17.33 ± 0.02 &
  20.32 ± 0.01 \\ \cmidrule(l){2-5} 
 &
  $\mathrm{CosSim}(\Bar{E}_0, \Bar{E}_i) \rightarrow \frac{1}{|\bar{E}_0 - \bar{E}_i|_2}$ &
  13.29 ± 0.87 &
  14.48 ± 0.45 &
  16.75 ± 0.34 \\ \cmidrule(l){2-5} 
 &
  No $\sigma(\cdot)$ &
  12.88 ± 0.59 &
  13.74 ± 0.48 &
  16.45 ± 0.26 \\ \cmidrule(l){2-5} 
\multirow{-4}{*}{RewardBench} &
  \cellcolor[HTML]{EDEDE9}Ours &
  \cellcolor[HTML]{EDEDE9}\textbf{12.72 ± 0.30} &
  \cellcolor[HTML]{EDEDE9}\textbf{12.84 ± 0.48} &
  \cellcolor[HTML]{EDEDE9}\textbf{16.35 ± 0.36} \\ \midrule
 &
  $\log(r_i) \rightarrow r_i$ &
  13.88 ± 0.01 &
  15.88 ± 0.01 &
  15.45 ± 0.01 \\ \cmidrule(l){2-5} 
 &
  $\mathrm{CosSim}(\Bar{E}_0, \Bar{E}_i) \rightarrow \frac{1}{|\bar{E}_0 - \bar{E}_i|_2}$ &
  16.27 ± 0.81 &
  15.55 ± 0.83 &
  11.90 ± 1.07 \\ \cmidrule(l){2-5} 
 &
  No $\sigma(\cdot)$ &
  9.53 ± 0.11 &
  13.65 ± 0.01 &
  12.58 ± 0.01 \\ \cmidrule(l){2-5} 
\multirow{-4}{*}{LLMBar} &
  \cellcolor[HTML]{EDEDE9}Ours &
  \cellcolor[HTML]{EDEDE9}\textbf{8.03 ± 0.39} &
  \cellcolor[HTML]{EDEDE9}\textbf{9.95 ± 0.30} &
  \cellcolor[HTML]{EDEDE9}\textbf{8.61 ± 0.41} \\ \bottomrule
\end{tabular}%
}
\end{table*}

%%%%%%%%%%%%%%%%%%%%%%%%%%%%%%%%%%%%%%%%%%%%%%%%%%%%%%%%%%%%%%%%%%%%%%%%%%%%%%%
%%%%%%%%%%%%%%%%%%%%%%%%%%%%%%%%%%%%%%%%%%%%%%%%%%%%%%%%%%%%%%%%%%%%%%%%%%%%%%%

\end{document}